\documentclass[format=acmsmall, screen]{acmart}

\usepackage{amsmath}
\usepackage{amsfonts,pifont}
\usepackage{graphicx}
\usepackage{textcomp}
\usepackage{amsthm}

\usepackage{setspace}
\usepackage{subfigure}
\usepackage{algorithmic}
\usepackage[ruled,vlined,boxed,linesnumbered]{algorithm2e}
\usepackage{bm}
\usepackage{url}
\usepackage{hyperref}
\usepackage{booktabs}
\usepackage{array}
\usepackage{multirow}
\usepackage{makecell}
\usepackage{tabularx}
\newcolumntype{C}[1]{>{\centering}p{#1}}
\usepackage{float}
\usepackage{stfloats}
\usepackage{xcolor}

\usepackage{forest}
\usetikzlibrary{arrows.meta,shapes,positioning,shadows,trees}
\tikzset{
    basic/.style  = {draw, text width=1.4cm, drop shadow, font=\footnotesize, rectangle},
    root/.style   = {basic, rounded corners=2pt, thin, align=center,
                     fill=gray!10,text width=1.1cm,},
    fnode/.style = {basic, thin, rounded corners=2pt, align=center, fill=blue!10,text width=1.4cm},
    snode/.style = {basic, thin,rounded corners=2pt, align=left, fill=green!10, text width=2.6cm},
    tnode/.style = {basic, thin, rounded corners=2pt, align=left, fill=yellow!10,text width=6.5cm},
    wnode/.style = {basic, thin, align=left, fill=pink!10, text width=6.5em},
    edge from parent/.style={draw=black, edge from parent fork right}
}

\usepackage{threeparttable}
\SetCommentSty{mycommfont}
\SetKwInput{KwInput}{Input}                
\SetKwInput{KwOutput}{Output} 

\allowdisplaybreaks[4]

\AtBeginDocument{%
  \providecommand\BibTeX{{%
    \normalfont B\kern-0.5em{\scshape i\kern-0.25em b}\kern-0.8em\TeX}}}

\begin{document}

\title{Diffusion Models for Reinforcement Learning: Foundations, Taxonomy, and Development}


\author{Changfu Xu}
\email{xuchangfu@jxufe.edu.cn}
\affiliation{%
  \institution{Jiangxi University of Finance and Economics}
  \city{Nanchang}
  \state{Jiangxi}
  \country{China}
  \postcode{330013}
}

\author{Jianxiong Guo}
\email{jianxiongguo@bnu.edu.cn}
\affiliation{%
  \institution{Beijing Normal University}
  \city{Zhuhai}
  \state{Guangdong}
  \country{China}
  \postcode{519087}
}

\author{Yuzhu Liang}
\email{cs\_yuzhuliang@163.com}
\affiliation{%
    \institution{Beijing Normal University}
    \city{Zhuhai}
    \state{Guangdong}
    \country{China}
    \postcode{519087}
}

\author{Haiyang Huang}
\email{younghuangh@163.com}
\affiliation{%
    \institution{Beijing Normal University}
    \city{Zhuhai}
    \state{Guangdong}
    \country{China}
    \postcode{519087}
}

\author{Haodong Zou}
\email{zou\_hd@163.com}
\affiliation{%
	\institution{Anhui University}
	\city{Hefei}
	\state{Anhui}
    \country{China}
	\postcode{230601}
}

\author{Xi Zheng}
\email{james.zheng@mq.edu.au}
\affiliation{%
  \institution{Macquarie University}
  \city{Sydney}
  \state{NSW}
  \country{Australia}
  \postcode{2109}
}

\author{Shui Yu}
\email{Shui.Yu@uts.edu.au}
\affiliation{%
  \institution{University of Technology Sydney}
  \city{Sydney}
  \state{NSW}
  \country{Australia}
  \postcode{2007}
}
\author{Xiaowen Chu}
\email{xwchu@hkust-gz.edu.cn}
\affiliation{%
  \institution{The University of Science and Technology (Guangzhou)}
  \city{Guangzhou}
  \state{Guangdong}
  \country{China}
  \postcode{511453}
}
\author{Jiannong Cao}
\email{csjcao@comp.polyu.edu.hk}
\affiliation{%
  \institution{The Hong Kong Polytechnic University}
  \country{Hong Kong}
  \postcode{999077}
}

\author{Tian Wang}
\email{tianwang@bnu.edu.cn}
\authornote{Corresponding author}
\affiliation{%
  \institution{Beijing Normal University}
  \city{Zhuhai}
  \state{Guangdong}
  \country{China}
  \postcode{519087}
}

\renewcommand{\shortauthors}{C. Xu et al.}

\begin{abstract}
    Diffusion Models (DMs), as a leading class of generative models, offer key advantages for reinforcement learning (RL), including multi-modal expressiveness, stable training, and trajectory-level planning. This survey delivers a comprehensive and up-to-date synthesis of diffusion-based RL. We first provide an overview of RL, highlighting its challenges, and then introduce the fundamental concepts of DMs, investigating how they are integrated into RL frameworks to address key challenges in this research field. We establish a dual-axis taxonomy that organizes the field along two orthogonal dimensions: a function-oriented taxonomy that clarifies the roles DMs play within the RL pipeline, and a technique-oriented taxonomy that situates implementations across online versus offline learning regimes. We also provide a comprehensive examination of this progression from single-agent to multi-agent domains, thereby forming several frameworks for DM-RL integration and highlighting their practical utility. Furthermore, we outline several categories of successful applications of diffusion-based RL across diverse domains, discuss open research issues of current methodologies, and highlight key directions for future research to advance the field. Finally, we summarize the survey to identify promising future development directions. We are actively maintaining a GitHub repository (\url{https://github.com/ChangfuXu/D4RL-FTD}) for papers and other related resources to apply DMs for RL.
\end{abstract}



\maketitle

\section{Introduction}\label{sec-1}
\subsection{Background}
\begin{figure}[!h]
    \centering
    \includegraphics[width=0.65\linewidth]{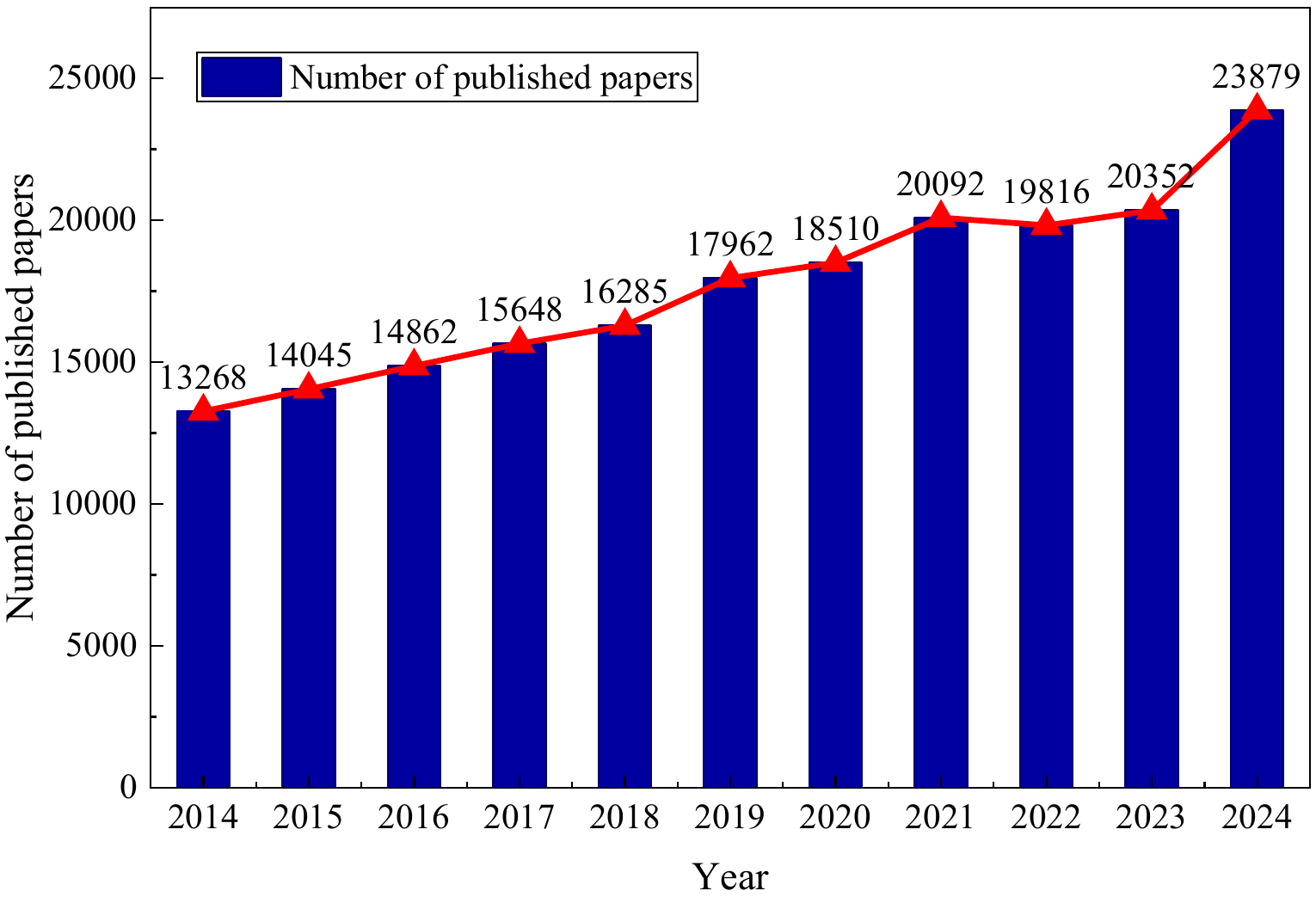}
    \caption{The number of published papers by searching ''Diffusion Model'' in Web of Science (Access date: June 23, 2025). Diffusion models have garnered significant attention in both research and industry fields, particularly over the last two years.}
    \label{number_of_published_papers}
\end{figure}

Diffusion Models (DMs) have recently emerged as a highly influential class of generative models, attracting widespread attention across the machine learning community \cite{esser2024scaling, diffusion2024platform}. Originally developed for high-quality data generation tasks such as image and video synthesis \cite{ho2020denoising}, DMs are generative denoising processes that learn to reverse a stepwise corruption of data, enabling the generation of realistic data samples. Compared to earlier generative approaches such as Variational Autoencoders (VAEs) \cite{kingma2014auto} and Generative Adversarial Networks (GANs) \cite{goodfellow2014generative}, DMs offer notable advantages in producing high-fidelity samples and ensuring greater training stability. As a result, the versatility and potency of DMs have been demonstrated in a broad range of application areas, as shown in Fig. \ref{number_of_published_papers}, including Computer Vision (CV) \cite{ho2020denoising, lugmayr2022repaint}, Natural Language Processing (NLP) \cite{austin2021structured, li2022diffusionlm}, audio generation \cite{lee2021nu, kong2021diffwave}, and particularly in sequential decision-making \cite{ajay2023conditional, huang2024diffusion, janner2022planning, du2024diffusion, xu2025enhancing}.

Reinforcement Learning (RL) has been widely applied to various domains, such as robot control \cite{de2022insect}, autonomous driving \cite{hu2023planning}, and task scheduling \cite{liang2024efficient}. The fundamental goal of RL is to learn a policy that maps observations or states to actions, thereby maximizing the cumulative reward over time. Furthermore, RL, combined with Deep Neural Networks (DNNs), has been formulated into Deep Reinforcement Learning (DRL). The power of DRL stems from its ability to learn directly from high-dimensional sensory inputs, such as images and raw sensor data, without requiring hand-engineered features. This end-to-end learning paradigm allows DRL agents to discover intricate patterns and optimal control policies that might be difficult or impossible to define manually. It includes two categories of methods: (1) value-based methods, including Deep Q-value Network (DQN) \cite{mnih2015human}, Double DQN \cite{van2016deep}, and Dueling DQN \cite{wang2016dueling}; and (2) policy-based methods based on the Actor-Critic framework, including stochastic policy gradient methods: Trust Region Policy Optimization (TRPO) \cite{schulman2015trust} and Proximal Policy Optimization (PPO) \cite{schulman2017proximal}, and deterministic policy gradient methods \cite{silver2014deterministic}: Deep Deterministic Policy Gradient (DDPG) \cite{lillicrap2015continuous}, Twin Delayed Deep Deterministic (TD3) \cite{fujimoto2018addressing}, and Soft Actor-Critic (SAC) \cite{haarnoja2018soft}. DRL has significantly broadened the scope of sequential decision-making, enabling its application to complex, large-scale problems. Despite these advancements, DRL methods still face several critical limitations. \textit{First}, DRL methods often suffer from sample inefficiency, as their policy learning frequently requires a large number of interactions with the environment. \textit{Second}, many DRL algorithms depend on stochastic sampling from unimodal distributions (e.g., Gaussians), which may inadequately capture complex or multimodal action spaces \cite{zhu2023diffusionrl}. \textit{Third}, several challenges, such as bootstrapping errors, off-policy learning complications, and high sensitivity to hyperparameters, can hinder the convergence of DRL models \cite{fujimoto2019benchmarking}. \textit{Fourth}, conventional RL approaches typically model policies as direct state-to-action mappings, which can oversimplify the decision-making process and restrict representational capacity.

\begin{figure}[!t]
    \centering
    \includegraphics[width=\linewidth]{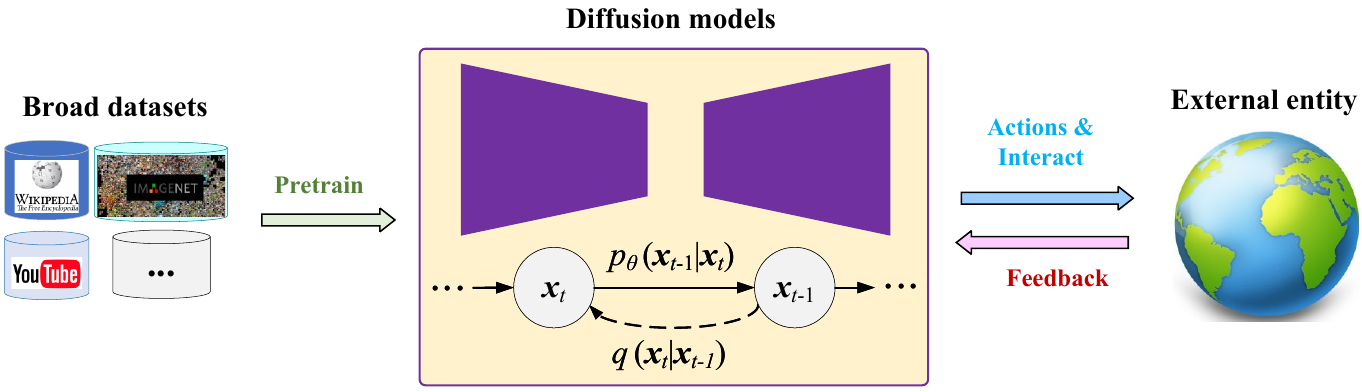}
    \caption{Overview of DMs for RL. DMs pretrained on broad data are adapted to accomplish specific tasks by making actions, interacting with external entities, and receiving feedback.}
    \label{overview_D4RL}
\end{figure}

\subsection{Motivation}
In response to these limitations, DMs have recently been applied to optimize RL techniques. They have shown strong performance in offline trajectory modeling, planning, and goal-conditioned control, with a particular focus on DRL, as illustrated in Fig. \ref{overview_D4RL}. A representative example is Diffuser \cite{janner2022planning}, which leverages a DM to learn trajectory distributions from offline datasets and performs goal-directed planning through guided sampling. Building on this foundation, numerous subsequent works have integrated DMs into various components of the RL pipeline \cite{lu2023contrastive, xu2024phd}. For instance, the DM is used to replace conventional Gaussian policies \cite{wang2023diffusion}, augment experience data \cite{lu2023synthetic}, or extract latent skill representations \cite{venkatraman2023reasoning}. These methods have demonstrated strong performance across a range of applications, including multitask RL \cite{he2023diffusion}, Imitation Learning (IL) \cite{hegde2023generating}, and trajectory generation \cite{carvalho2023motion}. Most notably, the expressive and flexible distribution modeling capacity of DMs offers promising solutions to several long-standing challenges in RL, such as effective exploration, policy expressiveness, and planning under uncertainty. Therefore, when these DMs are applied to RL, they can model the entire trajectory of states and actions as a sample from a learned distribution, rather than predicting actions one step at a time, offering several compelling advantages as follows:
\begin{itemize} 
    \item \textbf{Improved Exploration.} DMs can naturally represent complex, multimodal action or trajectory distributions to enable the generation of more diverse behavior, thus improving exploration \cite{janner2022planning}.
    \item \textbf{Trajectory-level Reasoning.} Instead of selecting actions incrementally, diffusion-based policies can generate full sequences conditioned on task objectives, enabling better long-term planning \cite{ajay2023conditional}.
    \item \textbf{Stability and Generalization.} The denoising diffusion process-based training paradigm often results in smoother optimization landscapes and better generalization, especially in offline RL settings \cite{fujimoto2019benchmarking}.
    \item \textbf{Compatibility with RL.} DMs can produce effective policies based on the training of fixed datasets, addressing the distribution shift and extrapolation errors that plague traditional RL algorithms \cite{kang2023efficient}.
\end{itemize}

As a result, a new class of diffusion-based decision-making models, such as Diffuser \cite{janner2022planning}, Decision Diffuser \cite{ajay2023conditional}, and Deep Diffusion-based SAC (D2SAC) \cite{du2024diffusion}, has emerged, showing strong empirical performance across a range of sequential decision-making benchmarks. These models blend the strengths of generative modeling with decision optimization, opening new avenues for robust, efficient, and expressive policy learning.

\begin{table*}[!b]
\caption{Summary of survey papers on DMs with different applications.}
\centering
\renewcommand{\arraystretch}{1.5}
\belowrulesep=0pt
\aboverulesep=0pt
\footnotesize
\begin{tabular}{m{30pt}<{}|m{210pt}<{}|m{110pt}<{}}
\toprule
\textbf{Survey} & \textbf{Contribution} & \textbf{Emphasis}\\
\midrule
\cite{cao2024survey} & Present the fundamental formulation of DMs, algorithmic enhancements, and the manifold applications of DMs & \multirow{3}{0.3\textwidth}{General review of DMs to provide advanced and comprehensive insights into diffusion and elucidate the DM's developmental trajectory and future directions.}\\
\cmidrule{1-2}
\cite{yang2023diffusion} & Provide a contextualized, in-depth look at the state of DMs, identifying the key areas of focus and pointing to potential areas for further exploration & \\
\midrule
\cite{croitoru2023diffusion} & Provide a comprehensive review of DMs applied in vision, comprising both theoretical and practical contributions in the field & \multirow{4}{0.3\textwidth}{The applications of DMs on CV, illustrating the current limitations of DMs and promising some interesting directions for future research.}\\
\cmidrule{1-2}
\cite{zhang2023text} & Discuss the DMs applied in image generation from text & \\
\cmidrule{1-2}
\cite{ulhaq2022efficient} & Present the most recent advances in DMs for vision from their computational efficiency viewpoint & \\
\midrule
\cite{zhu2023diffusionnlp} & Review the research results of DMs in the field of NLP from text generation, text-driven image generation, and other four aspects & \multirow{5}{0.3\textwidth}{The applications of DMs on NLP to investigate a comprehensive review of the use of DMs in NLP and explore further permutations of integrating Transformers into DMs}\\
\cmidrule{1-2}
\cite{zou2023survey} & Discusses the different formulations of DMs used in NLP, their strengths and limitations, and their applications & \\
\cmidrule{1-2}
\cite{li2023diffusion} & Review the recent progress in DMs for non-autoregressive text generation and the optimization techniques for text data & \\
\cmidrule{1-2}
\cite{zhang2023survey} & Provide the recent progress of diffusion-based speech synthesis & \\
\midrule
\cite{guo2023diffusion} & Provide an overview of the applications of DMs in bioinformatics to aid their further development in bioinformatics and computational biology & The applications of DMs in bioinformatics\\
\midrule
\cite{guo2023diffusion} & Present a survey on DMs for recommendation, and draw a bird's-eye view from the perspective of the whole pipeline in real-world recommender systems & The applications of DMs on recommender systems\\
\midrule
\cite{du2024enhancing} & Provide a comprehensive tutorial on the intelligent network optimization with DMs & The applications of DMs on network optimization\\
\bottomrule
\end{tabular}
\label{table-survey-summary}
\end{table*}

\subsection{Contribution}
Although several surveys on DMs exist as summarized in Table \ref{table-survey-summary}, they either offer broad overviews (e.g., \cite{cao2024survey} and Yang et al. \cite{yang2023diffusion}) or focus on specific domains such as CV (e.g., \cite{croitoru2023diffusion} and \cite{zhang2023text}) or NLP (e.g., \cite{zhu2023diffusionnlp} and \cite{zou2023survey}), leaving a gap in the comprehensive understanding of DMs for decision-making optimization. This survey seeks to bridge these gaps by systematically reviewing recent advances, categorizing key approaches, and highlighting existing challenges and future research directions. Our aim is to provide researchers and practitioners with a clear, structured, and up-to-date overview of this rapidly evolving field. Furthermore, while diffusion-based RL methods have also been explored in \cite{zhu2023diffusionrl}, they differ significantly in scope, methodology, and objectives. Specifically, compared with \cite{zhu2023diffusionrl}, this survey: 1) presents a comprehensive review of diffusion-based RL, covering the background, challenges, integration strategies, and current research trends in various domains, such as robotics, autonomous driving, edge computing, and more; 2) introduces a systematic taxonomy and structure, outlining representative works and applications; 3) looks into architectural and sampling issues specific to DMs in RL and discusses the value function roles of DMs in both single-agent RL and multi-agent RL, as well as online RL and offline RL from the perspective of technique taxonomy; 4) summarize recent advances, applications, and promotions of the diffusion-based RL methods in various fields; 5) highlights open issues and maps out future research prospects and emerging topics that are less thoroughly discussed in the earlier survey. The contributions of our survey are listed below:
\begin{itemize}
\item We provide a comprehensive tutorial on RL with DMs. This tutorial presents a comprehensive understanding of the origin, development, and key strengths of DMs, and outlines the role of DMs across various historical periods in RL techniques.
\item We classify the roles of DMs in RL into six categories: diffusion-based trajectory optimization, diffusion-based policy learning, diffusion-based IL, diffusion-based exploration augmentation, diffusion-based environmental simulation, and diffusion-based reward model. We also present the basic problem formulations of these six categories and depict the general solution framework for these six problems regarding the integration of DMs into RL. These studies provide a comprehensive review of the DMs for RL.
\item We have an in-depth look at DM for RL from a single-agent to multi-agent perspective, including both the DMs for single-agent RL and multi-agent RL from the perspective of function taxonomy, as well as the DMs for offline RL and online RL from the perspective of technique taxonomy, demonstrating the practicality and efficacy of the DMs for RL.
\item We summarize recent advances and promotions in RL with DMs for various applications, such as robotics, autonomous driving, and edge computing, and discuss potential directions for DMs research and applications, providing insights into how DMs can evolve and continue to influence future RL technology in modern applications.
\end{itemize}

\begin{figure}[!t]
    \centering
    \includegraphics[width=\linewidth]{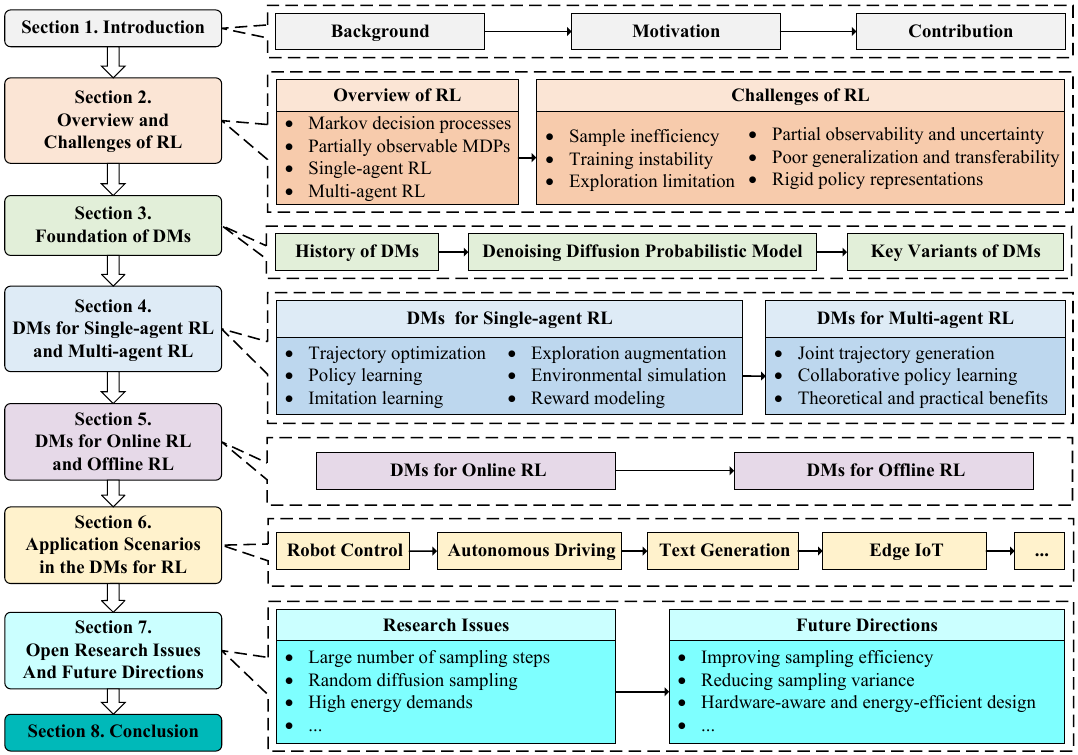}
    \caption{The taxonomy of this survey. We provide a comprehensive overview of the current research landscape at the intersection of DMs and RL by systematically analyzing this emerging field's progress, challenges, solutions, and opportunities.}
    \label{organization_structure}
\end{figure}

As shown in Fig. \ref{organization_structure}, the remainder of this survey is organized as follows: Section \ref{sec-RL-overview} first presents an overview of RL and then discusses the challenges in RL. Section \ref{diffusion_methodology} revisits the foundational concept of the DM and its key variants. Section \ref{function-taxonomy} discusses the DMs for single-agent RL and multi-agent RL from the perspective of function taxonomy. Section \ref{technique-taxonomy} further discusses DMs for online RL and offline RL from the perspective of technique taxonomy. Section \ref{sec-applications} describes recent advances and applications of the diffusion-based RL method. Section \ref{open-issues-directions} further analyzes open research issues and promising directions. The last Section \ref{conclusion} summarizes the survey with some concluding remarks.

\section{Overview and Challenges of RL}\label{sec-RL-overview}
In this section, we provide an overview of RL and briefly discuss several challenges of RL.

\subsection{Overview of RL}\label{SDM-overview}
In these settings, an agent must interact with an environment to achieve long-term objectives by executing a sequence of actions. This section provides an overview of key modeling frameworks, including Markov Decision Processes (MDPs), Partially Observable MDPs (POMDPs), and the major learning paradigms of single-agent RL and multi-agent RL.

\subsubsection{Markov Decision Processes}\label{MDP}
\begin{figure}[!t]
    \centering
    \includegraphics[width=0.6\linewidth]{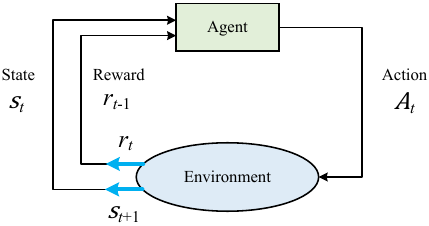}
    \caption{Framework of Markov decision process. By observing the current system state $\textbf{S}_{t}$ and reward $r_{t-1}$,the agent generates an action $\textbf{A}_{t}$. Then, by performing the $\textbf{A}_{t}$, the reward $r_{t}$ and next system state $\textbf{S}_{t+1}$ will be obtained from the environment. Finally, the $\textbf{S}_{t+1}$ and $\textbf{S}_{t+1}$ are inputted to the agent to generate the next action $\textbf{A}_{t+1}$.}
    \label{framework_markov_decision_making}
\end{figure}

The MDP is the standard framework for RL in fully observable stochastic environments, as shown in Fig. \ref{framework_markov_decision_making}. Formally, an MDP is defined by the tuple $(\mathcal{S}, \mathcal{A}, P, r, \gamma)$, where:
\begin{itemize}
    \item $\mathcal{S}$ is the state space,
    \item $\mathcal{A}$ is the action space,
    \item $P(s'|s,a)$ defines the state transition dynamics,
    \item $r(s,a)$ denotes the reward function,
    \item $\gamma \in [0,1)$ is the discount factor.
\end{itemize}

The goal is to learn a policy $\pi(a|s)$ that maximizes the expected cumulative discounted reward $\mathcal{J}(\pi)$:
\begin{equation}
    \mathcal{J}(\pi)=\mathbb{E}_{\pi} \left[ \sum\nolimits_{t=0}^{\infty} \gamma^t r(s_t, a_t) \right].
\end{equation}
Here, classical methods such as value iteration and policy iteration \cite{bellman1957dp, puterman2014mdp} can solve small-scale MDPs, but they struggle with scalability in large or continuous spaces.

\subsubsection{Partially Observable MDPs}
In real-world environments, agents often operate under uncertainty and cannot directly observe the true underlying state of the environment. This scenario is effectively modeled using POMDPs.

A POMDP is formally defined as a tuple:
\begin{equation}
\mathcal{M} = (\mathcal{S}, \mathcal{A}, \mathcal{O}, P, R, O, \gamma)
\end{equation}
where:
\begin{itemize}
    \item $\mathcal{S}$: Set of latent environment states.
    \item $\mathcal{A}$: Set of possible actions.
    \item $\mathcal{O}$: Set of possible observations received by the agent.
    \item $P(s'| s, a)$: State transition probability function, describing the probability of transitioning to state $s'$ given current state $s$ and action $a$.
    \item $r(s, a)$: Reward function, specifying the expected reward for taking action $a$ in state $s$.
    \item $O(o| s')$: Observation model, defining the probability of receiving observation $o$ in the system state $s'$.
    \item $\gamma \in [0, 1)$: Discount factor, representing the preference for immediate rewards over future rewards.
\end{itemize}

Unlike in fully observable MDPs, the agent in POMDPs does not have direct access to the state $s$, and must instead rely on partial observations $o \in \mathcal{O}$. To make decisions, the agent typically maintains a \textit{belief state} $b(s)$, which is a probability distribution over possible states, updated based on action-observation histories using Bayes' rule.

Due to the continuous and high-dimensional nature of the belief space, solving POMDPs exactly is computationally intractable in most practical settings. This motivates the use of RL to enable tractable decision-making under partial observability~\cite{kaelbling1998pomdp}. RL algorithms learn optimal policies via trial-and-error interaction with the environment, receiving feedback through rewards. Furthermore, RL can be categorized into single-agent RL and multi-agent RL approaches.

\subsubsection{Single-agent RL}
In the standard single-agent RL setting, the interaction between an agent and the environment can be modeled as an MDP, defined by a tuple $(\mathcal{S}, \mathcal{A}, P, r, \gamma)$ (see the above Section \ref{MDP}). 

Over the years, many algorithms have been proposed for single-agent RL, spanning from value-based methods to policy optimization techniques. The single-agent RL approach usually has two types: model-free and model-based. The model-free methods (e.g., DQN \cite{mnih2015human}, PPO \cite{schulman2017proximal}, SAC \cite{haarnoja2018soft}) learn value functions or policies directly. In contrast, the model-based methods attempt to learn a transition model and use planning techniques to derive policies \cite{moerland2023model}. Some representative approaches include:

\begin{itemize}
    \item \textbf{Q-learning} \cite{watkins1992q}: A model-free, off-policy algorithm that learns the optimal action-value function using temporal-difference updates.
    \item \textbf{DQN} \cite{mnih2015human}: Extends Q-learning using deep neural networks as function approximators and experience replay to enable learning from high-dimensional state spaces (e.g., images).
    \item \textbf{Policy Gradient (PG)} \cite{sutton1999policy}: Directly optimizes the policy by estimating gradients of expected returns, suitable for continuous action spaces.
    \item \textbf{PPO} \cite{schulman2017proximal}: A stable and efficient on-policy algorithm that clips policy updates to avoid large deviations, widely adopted in practice.
    \item \textbf{SAC} \cite{haarnoja2018soft}: A maximum entropy RL algorithm that trades off reward maximization and policy entropy, achieving strong performance in continuous control tasks.  
    \item \textbf{Actor-Critic Methods} \cite{konda1999actor}: Combines policy (actor) and value function (critic) learning to reduce variance in policy gradients while maintaining low bias.
\end{itemize}

These methods form the core of modern RL research and serve as baselines for extensions to more complex scenarios, such as hierarchical RL, offline RL, and multi-agent RL.

\subsubsection{Multi-agent RL}
Multi-agent RL extends the traditional RL framework to environments that involve multiple agents interacting with POMDPs. In multi-agent RL, the environment is modeled as a Markov Game (also known as a Stochastic Game), defined by a tuple $\mathcal{G} = (\mathcal{N}, \mathcal{S}, \{\mathcal{A}_i\}_{i \in \mathcal{N}}, P, \{R_i\}_{i \in \mathcal{N}}, \gamma)$, where:
\begin{itemize}
    \item $\mathcal{N} = \{1, 2, \dots, N\}$: the set of agents.
    \item $\mathcal{S}$: the shared state space.
    \item $\mathcal{A}_i$: the action space of agent $i$.
    \item $P(s'|s, \mathbf{a})$: the transition function with joint action $\mathbf{a} = (a_1, a_2, \dots, a_N)$.
    \item $R_i(s, \mathbf{a})$: the reward function for agent $i$.
    \item $\gamma \in [0,1)$: the discount factor.
\end{itemize}

Each agent $i$ aims to learn a policy $\pi_i(a_i|o_i)$ based on its observation $o_i$ to maximize its own expected cumulative reward:
\begin{equation}
    \mathcal{J}_i(\pi_i) = \mathbb{E}_{\pi_1, \dots, \pi_N} \left[ \sum\nolimits_{t=0}^{\infty} \gamma^t R_i(s_t, \mathbf{a}_t) \right].
\end{equation}
In such settings, each agent learns a policy to maximize its own expected cumulative reward, often in the presence of other learning agents. This interaction leads to non-stationarity from the perspective of any single agent, introducing challenges such as instability, policy co-adaptation, and partial observability. Multi-agent RL is widely applied in domains like autonomous driving, swarm robotics, smart grids, and collaborative control. Moreover, many representative multi-agent RL methods have been proposed to tackle these challenges:

\begin{itemize}
    \item \textbf{Independent Q-Learning (IQL)} \cite{tan1993multi}: Treats each agent as an independent learner using standard Q-learning, which may suffer from instability due to non-stationarity.
    \item \textbf{Multi-agent DDPG} \cite{lowe2017multi}: Utilizes centralized training with decentralized execution, where a centralized critic observes the global state and all agents’ actions to improve training stability.
    \item \textbf{Value Decomposition Networks (VDN)} \cite{sunehag2017value}: Decomposes a global joint action-value function into a summation of individual agent value functions, facilitating cooperative behavior learning. 
    \item \textbf{QMIX} \cite{rashid2018qmix}: Extends VDN by learning a monotonic mixing network to combine individual Q-values into a joint Q-value, enabling more expressive coordination mechanisms.
    \item \textbf{Multi-agent PPO} \cite{yu2022surprising}: Adapts the PPO framework to the multi-agent setting using a centralized critic for each agent, balancing exploration and stability.   
    \item \textbf{Hierarchical and Policy Factorization Methods} \cite{papoudakis2020benchmarking, he2024hierarchical}: Introduce structured policy representations and hierarchical learning to enable efficient credit assignment and long-horizon planning.
\end{itemize}

Recent multi-agent RL research trends include decentralized training, graph-based interaction modeling, multi-agent communication protocols, and the integration of large-scale pre-trained models for generalization. However, despite their effectiveness, RL methods often struggle with sample inefficiency and unstable convergence, particularly in complex environments or those with sparse rewards.

\subsection{Challenges of RL}
Sequential decision-making aims to find optimal strategies over time, typically formalized through frameworks such as MDPs and POMDPs. While MDPs assume full observability of the environment state, POMDPs extend the formulation to scenarios with uncertainty and incomplete observations. On top of these models, RL and IL offer complementary paradigms for policy acquisition. However, both traditional frameworks face critical challenges in real-world applications:

\subsubsection{Sample Inefficiency}
RL methods often require an extensive number of interactions with the environment to converge to satisfactory policies, particularly in high-dimensional or sparse-reward settings. This sample inefficiency limits their practicality in domains where interactions are costly or time-consuming, such as robotics, healthcare, or financial systems. IL methods can reduce the reliance on exploration by learning from expert demonstrations, yet they frequently struggle with generalization beyond the demonstration distribution, leading to compounding errors over time.

\subsubsection{Training Instability}
Many RL algorithms or models, such as DQN, PPO, and SAC, rely on complex optimization strategies involving off-policy learning, bootstrapping, and target networks. These mechanisms introduce non-stationarity in the dynamic learning process, making training highly sensitive to hyperparameters and prone to instability. For example, small variations in learning rate, exploration parameters, or reward scaling can also cause significant performance degradation or even learning collapse.

\subsubsection{Exploration Limitations}
Exploration strategies in RL often rely on simple stochastic perturbations (e.g., Gaussian noise or epsilon-greedy actions), which are insufficient in environments with deceptive rewards or multimodal action distributions. Consequently, agents may become trapped in local optima, failing to discover more optimal behaviors. This issue is exacerbated in environments with long time horizons or hierarchical task structures.

\subsubsection{Partial Observability and Uncertainty}
In many realistic scenarios, agents operate under partial observability, where only a subset of the true environmental state is accessible. While POMDPs provide a principled approach to modeling such settings, solving them is computationally intractable in most real-world applications. RL algorithms adapted for partial observability (e.g., those using recurrent policies) often suffer from performance degradation due to noisy observations and ambiguity in state estimation.

\subsubsection{Poor Generalization and Transferability}
Policies learned through RL or IL are typically tailored to the specific training environment and may not generalize well to unseen states or altered dynamics. This poses a significant barrier in real-world applications, where variations in environmental conditions are common. Moreover, models trained in simulation environments often experience a severe drop in performance when transferred to the real world.

\subsubsection{Rigid Policy Representations}
Conventional RL policies are usually parameterized as deterministic or stochastic mappings from states to actions. While effective in simple scenarios, these representations may lack the expressiveness required to capture diverse, multimodal behavior patterns. This rigidity hinders the ability to model uncertainty and to represent conditional strategies based on long-term planning or multiple behavioral modes.
\section{Foundations of DMs}\label{diffusion_methodology}
DMs, as a type of generative model, have recently gained significant traction due to their ability to generate high-fidelity samples in complex data domains such as images, audio, and trajectories. At their core, DMs operate by learning to reverse a gradually applied noising process, thereby transforming random noise into structured data. This section provides a concise overview of the formulation of DMs, including the original Denoising Diffusion Probabilistic Model (DDPM) \cite{ho2020denoising} and key variants that improve sampling efficiency and flexibility. 

\subsection{History of DMs}
\begin{figure}[!h]
    \centering
    \includegraphics[width=\linewidth]{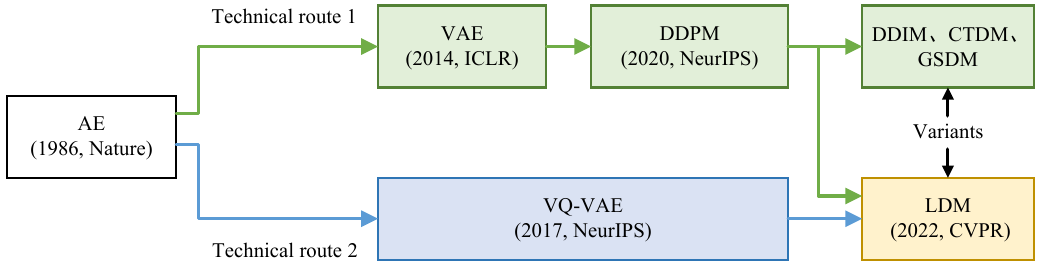}
    \caption{The technical roadmap of DM development. There are technical routes: 1) From Autoencoder (AE), Vector Autoencoder (VAE), and DDPM to DDIM, CTDM, and GSDM; 2) From AE and Vector Quantised-Variational Autoencoder (VQ-VAE) to Latent Diffusion Model (LDM).}
    \label{history_DMs}
\end{figure}

DMs, also known as DDPMs, have rapidly emerged as a powerful class of generative models. The foundational idea behind DMs is to learn the inverse of a gradual noising process, enabling the generation of complex data distributions from pure noise. This approach was first proposed by \cite{sohl2015deep}, who introduced a probabilistic framework that interprets data generation as a Markovian reverse diffusion process. The technical roadmap of DM development consists of two routes, as shown in Fig. \ref{history_DMs}.

\subsubsection{Technical Route 1} 
The DM development can be traced back to the Autoencoder (AE) proposed by Van \textit{et al.} in 1986 \cite{van2017neural}. AE is an early image generation model that utilizes the Encoder-Decoder architecture; however, it suffers from issues of overfitting and poor image reproduction quality. VAE \cite{kingma2014auto} introduces distribution learning on the basis of AE, which alleviates the overfitting phenomenon, but the generated image is still blurry. Subsequently, Ho \textit{et al.} \cite{ho2020denoising} propose the DDPM published on the NeurIPS conference in 2020 \cite{ho2020denoising}. The DDPM framework demonstrates that high-quality image synthesis could be achieved using a simple Gaussian noise schedule and a denoising network trained with a reweighted variational bound. This work sparks renewed interest in DMs and leads to substantial performance improvements across various modalities, becoming the foundation of DMs. However, DDPM is computationally expensive and slower to generate. Afterward, several improvements and variants of the original DDPM have been proposed to address issues related to sampling speed, flexibility, and continuous-time modeling. The most notable variants include Denoising Diffusion Implicit Models (DDIMs), Continuous-Time DMs (CTDMs), and Guided Sampling DMs (GSDMs) \cite{zhu2023diffusionrl}.

\subsubsection{Technical Route 2} 
On the other hand, Van \textit{et al.} \cite{van2017neural} propose a simple but powerful generative model, i.e., Vector Quantised-Variational AutoEncoder (VQ-VAE), to address the issue of AE overfitting. A significant contribution of the VQ-VAE is that it substantially reduces computational costs by compressing images into compact, discrete representations. Then, inspired by this advantage, the Latent Diffusion Model (LDM) \cite{rombach2022high} is proposed to further optimize generation efficiency and quality by combining the latent space compression of VQ-VAE with the diffusion denoising process of DDPM, achieving state-of-the-art performance in the field of image generation. LDM also introduces a conditioning module for multimodal information processing, making the generation process more flexible.

The growing interest in diffusion-based generative modeling for temporal and decision-making tasks has led to a new research frontier where DMs are being integrated into online and offline RL, IL, and multi-agent systems. This evolution marks a significant shift in how generative models contribute to solving complex planning and control problems.

\subsection{Denoising Diffusion Probabilistic Model}
\begin{figure*}[!h]
    \centering
    \includegraphics[width=\linewidth]{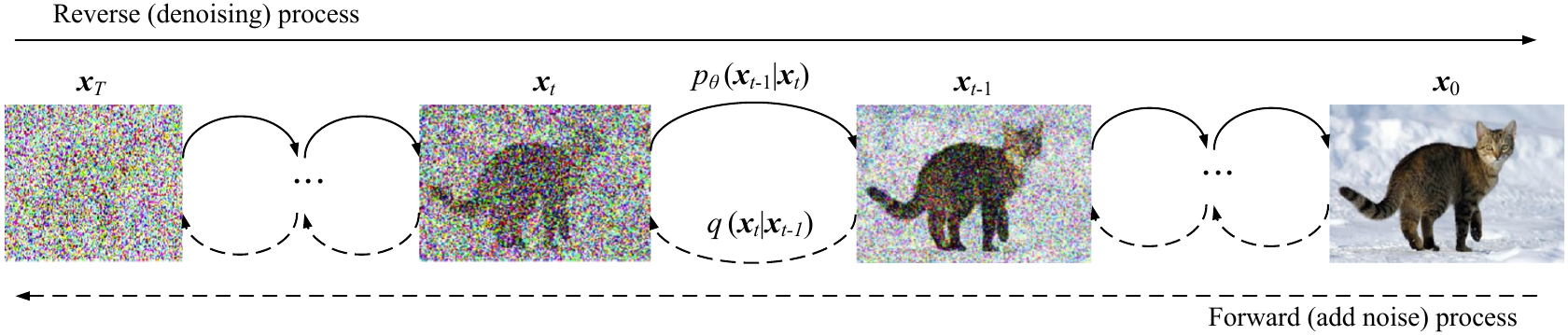}
    \caption{Progress of DMs. Given a target probability distribution $\mathbf{x}_{0}$, the forward process adds a sequence of Gaussian noises at each step to obtain $\mathbf{x}_{1}$, $\mathbf{x}_{2}$, ..., $\mathbf{x}_{T}$. The reverse process, also called the denoising process, infers the target $\mathbf{x}_{0}$ from a noise sample $\mathbf{x}_T \sim \mathcal{N}(\mathbf{0}, \mathbf{I})$ by removing noise.}
    \label{diffusion_framework}
\end{figure*}

The DDPM \cite{ho2020denoising} represents the standard formulation of DMs, using a fixed forward process and a learned reverse process for sample generation.  A DDPM typically consists of two components: a forward diffusion process that incrementally adds noise to input data, and a reverse denoising process that learns to invert this degradation. The basic progress of DMs is shown in Fig. \ref{diffusion_framework}.

\subsubsection{Forward Process}
In the forward process of the DM, a data point $\mathbf{x}_0 \sim q(\mathbf{x}_0)$ is gradually perturbed through a sequence of $T$ steps using a fixed Markov chain. At each step $t$, Gaussian noise is added according to a schedule:
\begin{equation}\label{forward-process}
q(\mathbf{x}_t | \mathbf{x}_{t-1}) = \mathcal{N}(\mathbf{x}_t; \sqrt{1 - \beta_t} \mathbf{x}_{t-1}, \beta_t \mathbf{I}),
\end{equation}
where $\beta_t \in (0, 1)$ is the noise variance at time step $t$. After $T$ steps, the data is effectively destroyed, and $\mathbf{x}_T \sim \mathcal{N}(\mathbf{0}, \mathbf{I})$ approximates standard Gaussian noise.

\subsubsection{Reverse Process}
The reverse process of the DM is parameterized by a neural network $\boldsymbol{\theta}$ that learns to reverse the diffusion and reconstruct $\mathbf{x}_0$ from noisy samples. The generative model defines a distribution:
\begin{equation}\label{reverse-process}
p_\theta(\mathbf{x}_{t-1} | \mathbf{x}_t) = \mathcal{N}(\mathbf{x}_{t-1}; \mu_\theta(\mathbf{x}_t, t), \Sigma_\theta(\mathbf{x}_t, t)),
\end{equation}
which is trained to approximate the true reverse process $q(\mathbf{x}_{t-1} | \mathbf{x}_t, \mathbf{x}_0)$. The network is typically trained to predict either the original data $\mathbf{x}_0$, the added noise $\boldsymbol{\epsilon}$, or the denoised sample using a simplified objective derived from the variational lower bound \cite{ho2020denoising}.

\subsection{Key Variants of DMs}
Although DDPMs achieve high-quality generation, they typically require hundreds or thousands of steps during inference, resulting in slow sampling. Afterwards, several improvements and variants of the original DDPM have been proposed to address issues related to sampling speed, flexibility, and continuous-time modeling. The most notable variants include Denoising Diffusion Implicit Models (DDIM) and Continuous-Time DMs (CTDMs).

\subsubsection{Denoising Diffusion Implicit Models} 
To address the inefficiency of DDPMs, DDIMs \cite{song2021denoising} introduce a non-Markovian deterministic sampling process that maintains sample quality while reducing the number of inference steps. The key idea is to define an implicit generative process using deterministic transformations:
\begin{align}
\mathbf{x}_{t-1} = & \sqrt{\alpha_{t-1}} \left( \frac{\mathbf{x}_t - \sqrt{1 - \alpha_t} \boldsymbol{\epsilon}_\theta(\mathbf{x}_t, t)}{\sqrt{\alpha_t}} \right) \nonumber \\
  & + \sqrt{1 - \alpha_{t-1}} \boldsymbol{\epsilon}_\theta(\mathbf{x}_t, t),
\end{align}
where $\alpha_t = \prod_{s=1}^t (1 - \beta_s)$. DDIM enables fast inference and preserves the flexibility to interpolate between stochastic and deterministic sampling.

\subsubsection{Continuous-Time Diffusion Models} CTDMs extend discrete diffusion steps to a continuous framework, treating the forward and reverse processes as solutions to Stochastic Differential Equations (SDEs) or Ordinary Differential Equations (ODEs) \cite{song2021score}. The forward SDE is defined as:
\begin{equation}
d\mathbf{x}(t) = f(\mathbf{x}(t), t) dt + g(t) d\mathbf{w}(t),
\end{equation}
where $\mathbf{w}(t)$ is the standard Wiener process. The reverse-time process $g(t)$ also follows an SDE that can be solved using learned score functions, i.e.,  $g(t) = \nabla_{\mathbf{x}} \log p_t(\mathbf{x})$. This formulation allows the use of score matching for training and provides a unified view of DMs as continuous generative flows. Key advantages of CTDMs include: adaptive step sizes for more efficient sampling, flexible conditioning mechanisms for controllable generation, and compatibility with advanced solvers (like DPM-Solver \cite{lu2022dpm}, which accelerates sampling with high precision.)

\subsubsection{Guided Sampling Diffusion Models}
GSDMs focus on the conditioned data distribution $p(\mathbf{x}|\mathbf{y})$ to generate samples with attributes of the label $\mathbf{y}$. According to whether an extra classifier model is adopted, the GSDM methods are divided into two categories: classifier GSDMs and classifier-free GSDMs. 

\textbf{Classifier GSDMs.} A key advantage of classifier GSDMs is that the classifier and the DM are trained independently. In other words, we just need to train a classifier to represent $p(\mathbf{y}|\mathbf{x})$ and then integrate it into an existing DM in sampling. In particular, we can train an extra classifier $p(\mathbf{y}|\mathbf{x}_t)$  based on noisy samples $\mathbf{x}_{t}$. Then, according to the pre-trained classifier and \cite{dhariwal2021diffusion}, the DM's reverse process is expressed as
\begin{align}
p_\theta(\mathbf{x}_{t-1} | \mathbf{x}_t, \mathbf{y}) = &\mathcal{N}(\mathbf{x}_{t-1}; \mu_\theta(\mathbf{x}_t, t) + \lambda \cdot \Sigma_\theta(\mathbf{x}_t,t)g(t), \nonumber \\
& \Sigma_\theta(\mathbf{x}_t, t)),
\end{align}
where $g(t) = \nabla_{\mathbf{x}} \log p_{\phi}(\mathbf{y}|\mathbf{x}_{t})$ and $\lambda$ represents a factor of the guidance scale.

\textbf{Classifier-free GSDMs.} Unlike the classifier GSDMs, classifier-free GSDMs should retrain the network model totally since their original training setups are modified. Thus, the classifier-free GSDMs are more expressive in training while achieving better performance. The classifier-free GSDMs aim to predict the score function $\nabla_{\mathbf{x}} \log p_t(\mathbf{x}|\mathbf{y})$. Through the Bayes Theorem, the score function can be represented an unconditional term and a classifier condition term, i.e., 
\begin{equation}\label{score-function}
    \nabla_{\mathbf{x}} \log p(\mathbf{x}|\mathbf{y}) = \nabla_{\mathbf{x}} \log p(\mathbf{y}|\mathbf{x}) + \nabla_{\mathbf{x}} \log p(\mathbf{x}).
\end{equation}
Furthermore, Song et al. \cite{song2021denoising} demonstrate that the DM and the score function are equivalent, meaning $\nabla_{\mathbf{x}_{t}} \log p(\mathbf{x}_{t}) \propto \boldsymbol{\epsilon}_\theta(\mathbf{x}_t, t)$. As a result, by substituting $\nabla_{\mathbf{x}} \log p(\mathbf{y}|\mathbf{x})$ with $\boldsymbol{\epsilon}_\theta(\mathbf{x}_t, y)$ into the equation (\ref{score-function}) and inferring, we have 
\begin{equation}\label{classifier-free}
\bar{\boldsymbol{\epsilon}}_w(\mathbf{x}_t, \mathbf{y}) = \boldsymbol{\epsilon}_\theta(\mathbf{x}_t, \mathbf{y}) + w \cdot ( \boldsymbol{\epsilon}_\theta(\mathbf{x}_t, \mathbf{y}) -\boldsymbol{\epsilon}_\theta(\mathbf{x}_t)),
\end{equation}
where $w$ represents the guidance scale. Finally, the equation (\ref{classifier-free}) is used for classifier guidance in classifier-free GSMDs.

\subsubsection{Latent Diffusion Models}
A limitation of standard DMs is their high computational cost when applied directly in the input space (e.g., pixels or state-action vectors). LDMs address this by performing diffusion in a compressed latent space, obtained through an autoencoder. Given an encoder $E$ and decoder $D$, data $\mathbf{x}_0$ is mapped into a latent representation:
\begin{equation}
    \mathbf{z}_0 = E(\mathbf{x}_0).
\end{equation}
The diffusion process is then applied in the latent space:
\begin{equation}
    q(\mathbf{z}_t \mid \mathbf{z}_{t-1}) = \mathcal{N}\!\left(\mathbf{z}_t; \sqrt{1 - \beta_t}\,\mathbf{z}_{t-1}, \, \beta_t \mathbf{I}\right).
\end{equation}
The reverse process is parameterized as:
\begin{equation}
    p_\theta(\mathbf{z}_{t-1} \mid \mathbf{z}_t) = \mathcal{N}\!\left(\mathbf{z}_{t-1}; \mu_\theta(\mathbf{z}_t, t, c), \, \Sigma_\theta(\mathbf{z}_t, t, c)\right),
\end{equation}
where $c$ denotes optional conditioning information, such as goals, rewards, or environment states in decision-making tasks. After denoising, the decoder reconstructs the data:
\begin{equation}
    \hat{\mathbf{x}}_0 = D(\mathbf{z}_0).
\end{equation}
LDMs offer two major benefits: \textbf{1) Efficiency}: Diffusion in a low-dimensional latent space significantly reduces computational cost. \textbf{2) Flexibility}: Conditioning mechanisms (e.g., cross-attention with goals or state inputs) enable controllable generation of trajectories, actions, or subgoals in sequential decision-making.

\subsection{DMs' Advantages for RL}
DMs have recently emerged as a powerful generative paradigm for modeling complex, multi-modal distributions \cite{zhu2023diffusionrl}. We observed that DMs provide a principled and expressive framework for generative modeling, characterized by a gradual noising process and its learned reversal. Therefore, DMs have emerged as a promising new direction to overcome the challenges in RL. By modeling trajectory distributions through a denoising generative process, they provide:
\begin{itemize}
    \item \textbf{Expressive policy representations} that capture multimodal and long-horizon behaviors.
    \item \textbf{Improved generalization} through generative pretraining and data augmentation.
    \item \textbf{Stable training dynamics} compared to adversarial and bootstrap-based methods.
\end{itemize}
These advantages position DMs as a powerful framework to advance the state-of-the-art in sequential decision-making under complex, dynamic, and uncertain environments. 

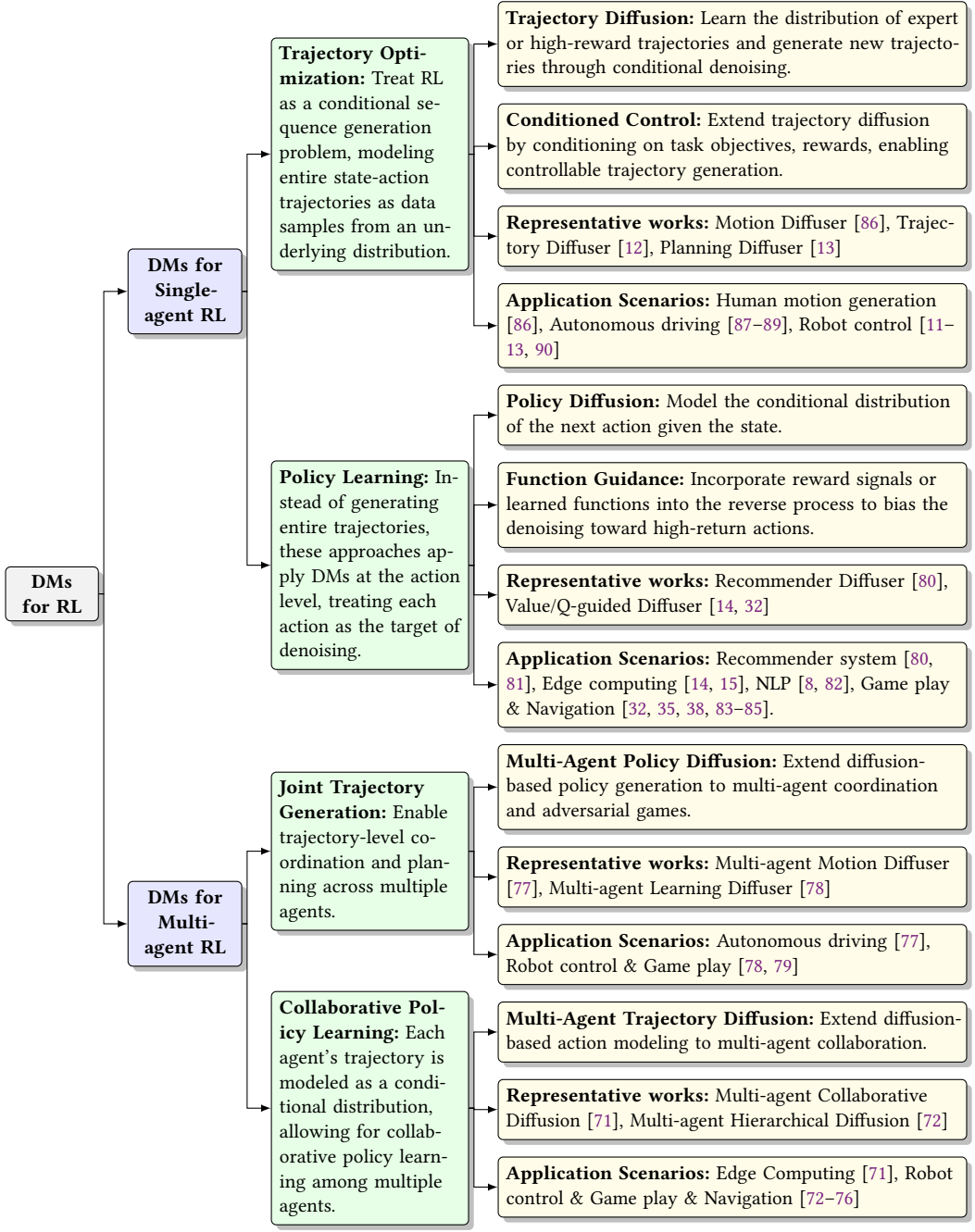
\begin{figure*}[t]
\centering
\begin{forest} for tree={
    grow=east,
    growth parent anchor=east,
    parent anchor=east,
    child anchor=west,
    edge path={\noexpand\path[\forestoption{edge},->, >={latex}] 
         (!u.parent anchor) -- +(2pt,0pt) |- (.child anchor)
         \forestoption{edge label};}
    }
[\textbf{DMs for RL}, root
    [\textbf{DMs for Multi-agent RL}, fnode
        [{\textbf{Collaborative Policy Learning:} Each agent’s trajectory is modeled as a conditional distribution, allowing for collaborative policy learning among multiple agents.}, snode
            [{\textbf{Application Scenarios:} Edge Computing \cite{yao2025enhancing}, Robot control \& Game play \& Navigation \cite{li2023hierarchical, wu2024diffusion, li2023beyond, qi2024diffusion, xu2024beyond}}, tnode]
            [{\textbf{Representative works:} Multi-agent Collaborative Diffusion \cite{yao2025enhancing}, Multi-agent Hierarchical Diffusion \cite{li2023hierarchical}}, tnode]
            [{\textbf{Multi-Agent Trajectory Diffusion:} Extend diffusion-based action modeling to multi-agent collaboration.}, tnode, yshift=-2.5cm]
        ]
        [{\textbf{Joint Trajectory Generation:} Enable trajectory-level coordination and planning across multiple agents.}, snode      
            [{\textbf{Application Scenarios:} Autonomous driving \cite{jiang2023motiondiffuser}, Robot control \& Game play \cite{zhu2024madiff, li2025dof}}, tnode]
            [{\textbf{Representative works:} Multi-agent Motion Diffuser \cite{jiang2023motiondiffuser}, Multi-agent Learning Diffuser \cite{zhu2024madiff}}, tnode]
            [{\textbf{Multi-Agent Policy Diffusion:}  Extend diffusion-based policy generation to multi-agent coordination and adversarial games.}, tnode, yshift=-1.8cm]
        ]
    ]
    [\textbf{DMs for Single-agent RL}, fnode
        [{\textbf{Policy Learning:} Instead of generating entire trajectories, these approaches apply DMs at the action level, treating each action as the target of denoising.}, snode
            [{\textbf{Application Scenarios:} Recommender system \cite{wang2023diffusionRM, yang2023generate}, Edge computing \cite{du2024diffusion, xu2025enhancing}, NLP \cite{li2022diffusionlm, gong2023diffuseq}, Game play \& Navigation \cite{zhang2024offline, he2023diffcps, kang2023efficient, hu2023instructed, he2023diffusion, wang2023diffusion}.}, tnode]
            [{\textbf{Representative works:} Recommender Diffuser \cite{wang2023diffusionRM}, Value/Q-guided Diffuser \cite{du2024diffusion, wang2023diffusion}}, tnode]
            [{\textbf{Function Guidance:} Incorporate reward signals or learned functions into the reverse process to bias the denoising toward high-return actions.}, tnode]
            [{\textbf{Policy Diffusion:} Model the conditional distribution of the next action given the state.}, tnode, yshift=-1.5cm]
        ]
        [{\textbf{Trajectory Optimization:} Treat RL as a conditional sequence generation problem, modeling entire state-action trajectories as data samples from an underlying distribution.}, snode            
            [{\textbf{Application Scenarios:} Human motion generation \cite{zhang2024motiondiffuse}, Autonomous driving \cite{wang2025diffad, chen2023polydiffuse, liao2025diffusiondrive}, Robot control \cite{ajay2023conditional, huang2024diffusion, janner2022planning, liang2023adaptdiffuser}}, tnode]
            [{\textbf{Representative works:} Motion Diffuser \cite{zhang2024motiondiffuse}, Trajectory Diffuser \cite{huang2024diffusion}, Planning Diffuser \cite{janner2022planning}}, tnode]
            [{\textbf{Conditioned Control:} Extend trajectory diffusion by conditioning on task objectives, rewards, enabling controllable trajectory generation.}, tnode]
            [{\textbf{Trajectory Diffusion:} Learn the distribution of expert or high-reward trajectories and generate new trajectories through conditional denoising.}, tnode, yshift=-3.0cm]
        ]
    ]
]
\end{forest}
\caption{Taxonomy of DMs for RL. Only the representative papers for each type of task are listed.}
\label{taxonomy-tree}
\end{figure*}

\begin{table*}[!t]
\caption{Summary of representative papers on DMs for single-agent RL.}
\centering
\renewcommand{\arraystretch}{1.5}
\belowrulesep=0pt
\aboverulesep=0pt
\footnotesize
\begin{tabular}{m{42pt}<{}|m{170pt}<{}|m{148pt}<{}}
\toprule
\textbf{Paper} & \textbf{Key Contribution} & \textbf{The Role of DMs}\\
\midrule
    ICML'22\cite{janner2022planning} & Introduce a denoising DM designed for trajectory data and behavior synthesis. & Employ a DM to refine trajectories iteratively.\\
    \midrule
    ICLR'22\cite{ajay2023conditional} & Investigate the conditional generative model to solve sequential decision-making directly. & Use a DM as the conditional generative model policy.\\
    \midrule
    arXiv'25\cite{wang2025diffad} & Present an end-to-end paradigm for autonomous driving based on DMs. & Leverage a DM to learn the latent distribution of bird’s-eye view images. \\
    \midrule 
    NIPS'23\cite{chen2023polydiffuse} & Propose a novel structured reconstruction algorithm for transforming visual sensor data into polygonal shapes with DMs  & Apply the forward diffusion process to train guidance networks, and use the reverse process to reconstruct polygonal shapes.\\
    \midrule
    CVPR'25\cite{liao2025diffusiondrive} & Introduce DMs into end-to-end autonomous driving to address the issues of mode collapse and heavy computational overhead. & Employ DMs to interact with conditional information in a cascaded manner, enabling more accurate trajectory reconstruction.\\
    \midrule
    TPAMI'24\cite{zhang2024motiondiffuse} & Propose a framework for applying DMs to text-driven human motion generation. & Utilize a DM to generate human motions through a series of denoising steps.\\
    \midrule
    TII'24\cite{zhang2024offline}, TMC'24\cite{du2024diffusion}, TMC'25\cite{xu2025enhancing} & Design a novel DRL method for improving behavior policy using DMs. & Train a reverse diffusion guide policy to generate the optimal action.\\    
    \midrule
    arXiv'23\cite{he2023diffcps} & Present a diffusion-based constrained policy search approach for offline RL. & Apply DMs to solve the limited expressivity problem of unimodal Gaussian policies.\\ 
    \midrule
    arXiv'23\cite{hu2023instructed} & Propose an effective conditional DM, referred to as the temporally-composable diffuser. & Integrate the DRL with a DM to extract temporal information from interaction sequences.\\ 
    \midrule
    NIPS'23\cite{he2023diffusion} & Propose a diffusion-based method for multi-task RL. & Incorporate a DM into transformer backbones for generative planning in multi-task RL\\ 
    \midrule
    NIPS'22\cite{li2022diffusionlm} & Develop a novel non-autoregressive language model for complex and controllable generation tasks with continuous DMs. & Leverage DMs to denoise a sequence of Gaussian vectors into word vectors, producing a series of intermediate latent variables.\\ 
    \midrule
    ICLR'23\cite{gong2023diffuseq} & Propose a DM designed for sequence-to-sequence text generation tasks. & Use the DM as a conditional language model to generate text.\\ 
    \midrule
    NIPS'23\cite{yang2023generate} & Reshape sequential recommendation as a learning-to-generate paradigm via a guided DM, achieving the user's true preference. & Generate an oracle item to reconstruct the positive item through the DMs' denoising process.\\ 
    \midrule
    SIGIR'23\cite{wang2023diffusionRM} & Present a novel diffusion recommender model to learn the generative process in a denoising manner, pointing out a promising future direction for generative recommender models. & Utilize DMs to perform high-dimensional categorical prediction and capture the time-sensitive dynamics of interaction sequences.\\ 
\bottomrule
\end{tabular}
\label{table-single-agent-RL-summary}
\end{table*}

\section{DMs for Single-agent RL and Multi-agent RL: Function Taxonomy }\label{function-taxonomy}
This section examines the current development of the DMS for RL from the perspective of function taxonomy. First, we have two categories: DMs for single-agent and DMs for multi-agent RL according to the number of agents in RL. We then present the taxonomy for these two categories, respectively, from the function of DMs in RL. Finally, we investigate the specific research status for each subcategory. The summary of representative papers on DMs for RL is given in the Fig. \ref{taxonomy-tree}. 

\subsection{DMs for Single-agent RL}\label{single-agent methodology}
In single-agent RL, DMs have emerged as a novel paradigm to improve policy expressiveness, trajectory optimization, and goal-conditioned behavior. Traditional RL methods typically assume unimodal policy distributions (e.g., Gaussians), which are limited in capturing complex, multimodal, or stochastic behaviors often needed in real-world environments. DMs address these limitations by learning generative models over trajectories or action distributions using a denoising process, as shown in Table \ref{table-single-agent-RL-summary}. 

\subsubsection{Diffusion-based Trajectory Optimization}
\begin{figure}[!t]
    \centering
    \includegraphics[width=0.8\linewidth]{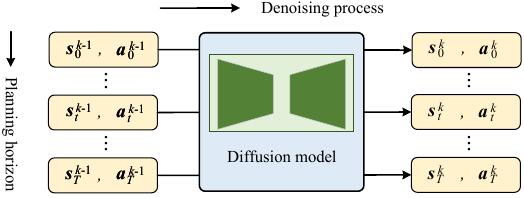}
    \caption{Framework of trajectory planning with diffusion-based single-agent RL. DMs are employed to generate trajectories, where the sampling objective focuses on portions of trajectories that may vary depending on the specific task requirements.}
    \label{diffusion_trajectory_planning}
\end{figure}

DMs can be used as planners to generate entire trajectories by modeling the joint distribution over sequences of states and actions \cite{wang2025diffad, zhang2024motiondiffuse, liang2023adaptdiffuser}. One representative method is \textit{Diffuser}, a pioneering framework that applies DDPMs to offline RL \cite{janner2022planning}. The model is trained on expert demonstrations, learning to reconstruct clean trajectory sequences from corrupted versions via a denoising process. During inference, trajectory optimization is conducted by conditioning the generation on desired outcomes, such as specific terminal states or cumulative rewards, thereby steering the planner toward task-specific goals. The general framework of diffusion-based RL solution for trajectory planning is illustrated in Fig.~\ref{diffusion_trajectory_planning}, highlighting the model’s ability to learn, generate, and adapt trajectories in a structured and goal-consistent manner. Let a trajectory be defined as $\tau = \{(s_1, a_1), (s_2, a_2), ..., (s_T, a_T)\}$. The forward process gradually adds noise to the trajectory:
\begin{equation}
    q(\tau_t | \tau_{t-1}) = \mathcal{N}(\sqrt{1 - \beta_t} \tau_{t-1}, \beta_t I),
\end{equation}
where $\beta_t$ is a variance schedule controlling the noise. The reverse process models the denoising dynamics:
\begin{equation}
    p_\theta(\tau_{t-1} | \tau_t) = \mathcal{N}(\mu_\theta(\tau_t, t), \Sigma_\theta(\tau_t, t)).
\end{equation}

During inference, the model samples trajectories conditioned on future constraints (e.g., desired goal states $s_T^*$), thus acting as a goal-conditioned planner. This approach bypasses the need for explicit dynamics modeling and provides implicit planning via denoising.

The implicit planning strategy of DMs offers a compelling alternative to conventional model-based planning methods, which often require explicit environment dynamics modeling and suffer from compounding errors in long-horizon predictions \cite{song2022learning}. By contrast, the diffusion-based method inherently captures complex temporal dependencies in trajectories and remains robust to multimodal behavior patterns \cite{jiang2023motiondiffuser, chen2023polydiffuse}. Furthermore, the flexibility to guide trajectory generation using conditional signals (e.g., goals, cost functions, constraints) empowers the framework to support diverse goal-directed behaviors that are typically challenging for traditional policy gradient or value-based approaches to learn directly. In addition to sample efficiency, this method exhibits strong generalization to out-of-distribution goals and tasks without the need for online environment interactions or fine-tuning. As such, it provides a unified, data-driven framework for planning in high-dimensional continuous control tasks.

\subsubsection{Diffusion-based Policy Learning}
\begin{figure}[!t]
    \centering
    \includegraphics[width=0.9\linewidth]{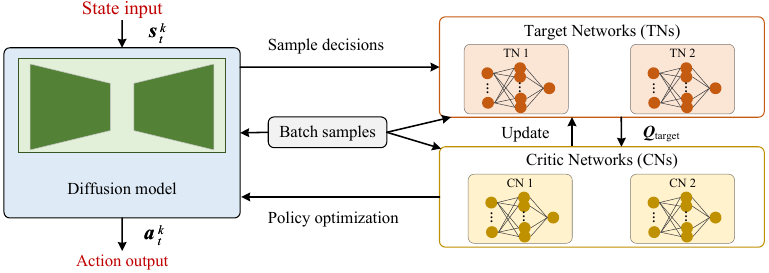}
    \caption{Framework of policy learning with diffusion-based single-agent RL. DMs are used to model policies by sampling actions conditioned on given states. The sampling process is typically guided either by target and critic networks using policy gradient-like methods or by directly incorporating guidance terms into the training objective.}
    \label{diffusion_policy_learning}
\end{figure}

Rather than optimizing over full trajectories, some recent works propose using DMs to represent stochastic policies directly, allowing actions to be sampled one step at a time based on the current observation \cite{wang2023diffusion, xu2025enhancing, xu2024accelerating, hansen2023idql}. The framework of diffusion-based policy learning in single-agent RL is illustrated in Fig.~\ref{diffusion_policy_learning}. This approach defines a denoising process in the action space, conditioned on the current state:
\begin{equation}
    \quad a_t^{(0)} = \text{ReverseDiffusion}_\theta(a_t^{(T)} | s_t),
\end{equation}
where $a_t^{(T)} \sim \mathcal{N}(0, I)$. In this setup, a DM is trained to generate an action $a_t$ conditioned on the current state $s_t$, by learning to reverse a noise-injection process, i.e., $a_t \sim p(a_t | s_t)$. This process effectively enables the model to sample from complex, high-dimensional, and multimodal action distributions that are difficult to capture with traditional unimodal Gaussian policies commonly used in RL.

By enabling a more expressive policy representation, diffusion-based policies support more diverse and adaptive behaviors across a wide range of tasks. For instance, Ajay et al. \cite{ajay2023conditional} introduced the \textit{Decision Diffuser} framework, which reinterprets decision-making as a conditional sequence modeling problem. The model generates sequences of actions and states conditioned on partial trajectories and desired outcomes. This formulation allows the agent to exhibit rich behavioral diversity and adapt to various task specifications or reward functions, without additional fine-tuning or retraining, making it highly suitable for offline RL and multi-objective optimization. Moreover, diffusion policies offer inherent flexibility in incorporating external constraints during sampling. Techniques such as classifier guidance \cite{dhariwal2021classifier} or classifier-free guidance enable conditioning the sampling trajectory on soft or hard constraints, including safety thresholds, energy budgets, or task-specific requirements. These mechanisms extend the usability of DMs in practical settings where dynamic constraints must be satisfied in real time. Overall, integrating DMs into policy learning introduces a powerful and general framework that not only improves expressiveness and robustness over conventional methods but also supports flexible control over the behavior generation process, aligning well with real-world decision-making scenarios.

\subsubsection{Diffusion-based IL}
IL \cite{ho2016generative} bypasses the need for rewards by leveraging expert demonstrations. The simplest form, Behavior Cloning (BC), treats the problem as supervised learning over state-action pairs:
\begin{equation}
  \pi_{\text{BC}} = \arg\min_{\pi}\frac{1}{|\mathcal{D}|}\sum\nolimits_{(s,a) \in \mathcal{D}}\ell(\pi(s),a),
\end{equation}
where $\mathcal{D}$ is a dataset of expert trajectories and $\ell(\cdot)$ is the loss. While I) is generally more sample-efficient than RL, it is susceptible to distributional shift, where small deviations from expert trajectories compound over time as the agent visits previously unseen states~\cite{jang2022bc}. For example, a robot learning to stack blocks from human demonstrations may execute slightly misaligned grasps early in a trajectory; without corrective feedback, these errors accumulate, causing failure in the later stages of the task.

To mitigate these issues, advanced approaches such as DAGGER \cite{ross2011dagger} incorporate corrective feedback by querying the expert on states visited by the learner. In the stacking robot example, DAGGER would provide corrective actions for misaligned grasps, reducing compounding errors. Inverse RL\cite{abbeel2004apprenticeship, ziebart2008maximum} offers another paradigm by inferring the underlying reward function from demonstrations, allowing the agent to generalize beyond the observed states. Similarly, generative adversarial IL \cite{ho2016generative} frames IL as a distribution-matching problem, encouraging the agent to generate behavior indistinguishable from the expert across the state-action space. These methods help IL handle longer horizons and complex behaviors, but still face challenges in modeling full trajectory distributions and handling multimodal behaviors.

DMs have recently been employed to overcome these limitations by modeling the full trajectory distribution, capturing multimodal and long-horizon behaviors more effectively. For example, Latent Diffusion Planning (LDP)~\cite{xie2025latent} improves imitation accuracy and efficiency, even when demonstrations are imperfect or suboptimal. In a robotic kitchen task, LDP can generate trajectories that correct slight errors in human demonstrations, such as adjusting grasp angles or timing motions, producing smoother and more reliable behavior.

SkillDiffuser~\cite{he2023diffusion} extends this idea by segmenting trajectories into reusable skills and modeling each skill with a diffusion process. For instance, in a pick-and-place task, a single “grasp object” skill learned via SkillDiffuser can be reused across multiple object types or locations, improving sample efficiency and generalization.

Diff-Control~\cite{liu2024diff} approaches IL from a state-space modeling perspective, using diffusion-based models to learn action representations that respect system dynamics. In applications like autonomous driving, this allows the model to capture the underlying structure of steering and acceleration behaviors over time, generating smooth and realistic vehicle trajectories that generalize better to novel road layouts or traffic conditions.

Collectively, these diffusion-based IL approaches provide more expressive policy representations, allowing agents to handle long-horizon tasks, complex or multimodal behaviors, and imperfect demonstration data. For example, in robotic manipulation, a diffusion-based IL model can imitate human stacking strategies while adjusting trajectories to prevent collisions, and in simulated driving, it can reproduce expert lane-change behaviors while adapting to new traffic patterns. By modeling full trajectory distributions rather than pointwise actions, DMs reduce compounding errors and improve robustness in imitation tasks, bridging a key gap in conventional IL approaches.

\subsubsection{Diffusion-based Exploration Augmentation}
Exploration remains a central challenge in RL, particularly in environments with sparse rewards, deceptive feedback, or high-dimensional action spaces. Standard exploration strategies, such as $\epsilon$-greedy exploration, entropy regularization \cite{mnih2016asynchronous, haarnoja2018soft}, and intrinsic motivation \cite{pathak2017curiosity, burda2018exploration}, often fail to generate sufficiently diverse or informative trajectories in such settings. Recently, DMs have been investigated as exploration augmenters, where they serve as powerful synthesizers of trajectories or state-action sequences, enabling agents to explore beyond the support of the collected data. A diffusion-based exploration augmenter can be formalized as learning a generative model over trajectories $\tau = (s_0, a_0, \dots, s_T, a_T)$:
\begin{equation}
q(\tau) = \prod\nolimits_{t=0}^T q(s_t, a_t),
\end{equation}
where the forward diffusion process gradually corrupts trajectories into Gaussian noise,
\begin{equation}\label{forward-process1}
q(\tau_t | \tau_{t-1}) = \mathcal{N}\big(\sqrt{1-\beta_t} \, \tau_{t-1}, \, \beta_t I\big),
\end{equation}
and the reverse process, parameterized by $\theta$, reconstructs trajectories by iteratively denoising,
\begin{equation}
p_\theta(\tau_{t-1} | \tau_t, c) = \mathcal{N}\big(\mu_\theta(\tau_t, c, t), \, \Sigma_\theta(\tau_t, c, t)\big),
\end{equation}
where $c$ denotes optional conditioning variables, such as novelty, reward, or uncertainty. During training, the DM is optimized with the standard noise-prediction objective~\cite{ho2020denoising}:
\begin{equation}
\mathcal{L}_{\text{DM}} = \mathbb{E}_{\tau, \epsilon, t} \left[ \left\| \epsilon - \epsilon_\theta(\sqrt{\bar{\alpha}_t} \tau + \sqrt{1 - \bar{\alpha}_t} \epsilon, t, c) \right\|^2 \right].
\end{equation}
For exploration augmentation, synthetic trajectories $\hat{\tau} \sim p\_\theta(\tau | c)$ are sampled and incorporated into policy updates, either by adding them to the replay buffer $\mathcal{D}$ (i.e., $\mathcal{D} \leftarrow \mathcal{D} \cup \{ \hat{\tau} \}$) or by directly regularizing the policy objective to encourage consistency with diffusion-generated trajectories.

Unlike local action perturbations, diffusion synthesizers generate entire plausible trajectories, enabling more global and structured exploration. In sparse-reward manipulation tasks, such as stacking blocks, a diffusion synthesizer can generate multiple feasible multi-step grasp-and-place trajectories. These samples expose the agent to diverse strategies, mitigating compounding errors. In goal-directed maze environments, diffusion-based exploration can produce diverse path candidates (shortcuts, detours, obstacle avoidance), improving the likelihood of reaching distant goals. In collaborative edge computing, diffusion synthesizers can propose diverse scheduling strategies for task allocation~\cite{xu2024accelerating}, improving robustness to uncertain workloads. In recommendation systems, synthetic user interaction sequences generated via diffusion can simulate novel preferences beyond logged data, reducing cold-start problems.

Exploration can be further enhanced by classifier guidance or reward-guided sampling \cite{dhariwal2021diffusion, janner2022planning}, where the reverse denoising process is biased toward regions of interest. For example, in autonomous driving, generated trajectories can be guided to prioritize safety constraints. Similarly, consistency models \cite{song2023consistency} can accelerate sampling, making exploration augmentation feasible in high-frequency online RL settings. Therefore, DMs represent a promising new paradigm for exploration augmentation, enabling agents to leverage trajectory-level synthesis for diverse, structured, and goal-directed exploration. Their integration into online and offline RL opens new directions for addressing one of the most fundamental bottlenecks in decision-making systems.

\subsubsection{Diffusion-Based Environmental Simulation} \label{sec:simulation}
Environmental simulation is a cornerstone of model-based reinforcement learning, where an agent leverages a learned dynamics model to generate synthetic rollouts for policy improvement \cite{moerland2023model}. Traditional simulators typically learn a parametric model $\hat{p}(s_{t+1}, r_t | s_t, a_t)$, often represented by neural networks \cite{janner2019trust}, to predict the next state and reward. However, these models are prone to \emph{model bias} and compounding errors when unrolled over long horizons.

DMs offer a more expressive alternative by modeling entire trajectories or future sequences directly. Instead of learning one-step dynamics, DMs approximate the distribution:
\begin{equation}
p_\theta(\tau) = p_\theta(s_0,a_0, \dots, s_T,a_T),
\end{equation}
where $\tau$ denotes a trajectory of states and actions. A forward noising process is applied to trajectories by Eq. (\ref{forward-process1}). The denoising model learns the reverse process:
\begin{equation}
p_\theta(\tau_{t-1}|\tau_t) \approx q(\tau_{t-1}|\tau_t, \tau_0),
\end{equation}
allowing the reconstruction of trajectories $\tau_0$ from Gaussian noise. By conditioning the diffusion model on the current state and action, environmental simulation becomes:
\begin{equation}
\hat{p}_\theta(s_{t+1:T}, a_{t:T} | s_{0:t}, a_{0:t}) \approx p(s_{t+1:T}, a_{t:T} | s_{0:t}, a_{0:t}),
\end{equation}
where the generative process produces multi-step rollouts that preserve temporal coherence.

Several works have demonstrated the benefits of diffusion-based simulators in decision-making. \textit{Trajectory Diffuser} \cite{janner2022planning} learns trajectory distributions from offline datasets, enabling the synthesis of plausible rollouts conditioned on goals or rewards. \textit{Motion Diffuser} \cite{zhang2024motiondiffuse} applies diffusion to simulate realistic motion trajectories, supporting safe planning in robotics. \textit{REDI} \cite{zhou2024redi} adapts diffusion-based simulation to online RL, enabling agents to refine policies under non-stationary dynamics. \textit{DriveDreamer} \cite{wang2024drivedreamer} integrates large-scale diffusion world models for autonomous driving, generating consistent and interpretable environment trajectories.

Compared to conventional one-step models, diffusion-based simulators capture multimodality, uncertainty, and long-horizon dependencies, reducing error accumulation. For instance, in robotics, they can simulate diverse yet physically plausible trajectories, while in autonomous driving, they generate realistic traffic interactions. However, challenges remain: diffusion simulation is computationally expensive, and integrating safety constraints is non-trivial. Promising directions include accelerating inference with DDIM~\cite{song2021denoising} and DPM-Solver~\cite{lu2022dpm}, or combining diffusion with latent variable models for lightweight rollout generation.

\subsubsection{Diffusion-Based Reward Modeling}
\label{sec:reward}
Reward modeling is essential in reinforcement learning (RL) for guiding policy optimization. Conventional approaches assume a scalar reward function $r(s,a)$ or learn it via supervised regression from demonstrations \cite{ziebart2008maximum, abbeel2004apprenticeship}. However, in real-world tasks, rewards may be \emph{implicit}, \emph{noisy}, or \emph{multimodal}, making them difficult to specify explicitly. DMs provide a flexible generative framework to capture such complexity in reward signals.

Let $\mathcal{D} = \{ (s_t,a_t,r_t) \}_{t=1}^N$ denote a dataset of transitions. Instead of directly learning a deterministic mapping $r_\phi: (s,a) \mapsto \hat{r}$, diffusion-based reward models learn a generative distribution over rewards conditioned on trajectories:
\begin{equation}
p_\theta(r_{0:T} | \tau) = p_\theta(r_0, \dots, r_T | s_0,a_0,\dots,s_T,a_T).
\end{equation}
The forward process perturbs the reward sequence:
\begin{equation}
q(r_t | r_{t-1}) = \mathcal{N}\big(\sqrt{1-\beta_t} r_{t-1}, \beta_t I\big),
\end{equation}
while the denoising model reconstructs clean rewards by learning:
\begin{equation}
p_\theta(r_{t-1}|r_t,\tau) \approx q(r_{t-1}|r_t,r_0).
\end{equation}
At inference, reward signals are generated as:
\begin{equation}
\hat{r}_{0:T} \sim p_\theta(r_{0:T}|\tau, c),
\end{equation}
where $c$ can encode task-specific conditions such as subgoals, safety constraints, or user preferences.

Diffusion-based reward modeling has been applied in several recent works. \textit{Decision Diffuser} \cite{ajay2023conditional} treats decision-making as conditional sequence modeling, where rewards guide trajectory denoising. \textit{MetaDiffuser} \cite{ni2023metadiffuser} incorporates classifier-guided sampling to generate reward-consistent trajectories for meta-RL tasks. \textit{AdaptDiffuser} \cite{liang2023adaptdiffuser} adapts diffusion reward models to changing environments, enabling robust performance under non-stationarity. \textit{Safe Reward Diffusion} \cite{wang2024safe} integrates constrained MDPs into the diffusion reward modeling framework, producing safety-aware reward shaping.

Compared to classical regression-based reward models, diffusion-based reward modeling can capture multimodal reward distributions, model long-horizon reward structures jointly with state-action sequences, and support conditioning on constraints, goals, or user preferences. However, training reward diffusion models can be computationally expensive, and their stochastic nature may introduce variance in policy optimization. Future directions include variance reduction strategies~\cite{song2023consistency}, integrating reward diffusion into online RL~\cite{zhou2024redi}, and combining with inverse RL frameworks~\cite{ho2016generative} for preference learning at scale.

\begin{table*}[!t]
\caption{Summary of representative papers on DMs for multi-agent RL.}
\centering
\renewcommand{\arraystretch}{1.5}
\belowrulesep=0pt
\aboverulesep=0pt
\footnotesize 
\begin{tabular}{m{40pt}<{}|m{170pt}<{}|m{145pt}<{}}
\toprule
\textbf{Paper} & \textbf{Key Contribution} & \textbf{The Role of DMs}\\
\midrule
    NIPS'24\cite{zhu2024madiff} & Propose a diffusion-based multi-agent learning framework to unify decentralized policy, centralized controller, teammate modeling, and trajectory prediction. & Realize with an attention-based DM to model the complex coordination among the behaviors of multiple agents.\\
    \midrule
    ICLR'25\cite{li2025dof} & Propose a DM-based offline multi-agent RL framework, significantly boosting the offline multi-agent RL algorithm compared to the original dataset. & Integrate the Q function directly into the DM as guidance to maximize the global returns, eliminating the need for separate training.\\
    \midrule
    CVPR'23\cite{jiang2023motiondiffuser} & Introduce a general and flexible framework for controlled and guided trajectory sampling, driven by arbitrary differentiable cost functions, enabling a wide range of novel applications. & Use conditional DMs to represent a permutation-invariant, multi-agent joint motion distribution.\\
    \midrule
    arXiv'23\cite{li2023beyond} & Present a novel diffusion offline multi-agent model that achieves significant improvements in performance, generalization, and data-efficiency. & Incorporate a DM into the policy network to enhance policy expressiveness and diversity.\\
    \midrule 
    APWCS'24\cite{qi2024diffusion} & Integrate DMs into the DRL technique to accelerate decision-making, considering communication among system agents. & Employ DMs to enhance multi-agent coordination, even when agents make decisions independently.\\
    \midrule
    arXiv'24\cite{xu2024beyond} & Propose global state inference for collaborative multi-agent RL with DMs. & Leverage DMs to reconstruct the global state using local observations. \\
    \midrule
    ICML'23\cite{li2023hierarchical} & Incorporate goals into the control-as-inference framework by formulating offline decision-making as a conditional generative modeling problem. & Use a reward-conditioned goal DM to discover subgoals, and a goal-conditioned trajectory DM to generate corresponding actions.\\
    \midrule
    arXiv'24\cite{wu2024diffusion} & Address motion planning for an evasive target in partially observable multi-agent adversarial pursuit and evasion games with DMs. & Integrate a DM to generate global paths that adapt to environmental observations.\\
    \midrule
    TSC'25\cite{yao2025enhancing} & Propose a novel vector database-assisted cloud-edge collaborative optimization framework of the quality of service of Large Language Models (LLMs), reducing response times for subsequent similar requests. & Employ a diffusion-based policy network to extract the request features of LLMs, determining whether to request the LLMs in the cloud or retrieve results from the edge.\\
\bottomrule
\end{tabular}
\label{table-multi-agent-DRL-summary}
\end{table*}

\subsubsection{Advantages Over Traditional Policies}
This paradigm of the DM for single-agent RL has demonstrated competitive or superior performance on robotic locomotion, manipulation, and navigation tasks compared to traditional DRL baselines. The specific advantages include:
\begin{itemize}
  \item \textbf{Multimodal action modeling:} DMs allow sampling diverse action modes, enhancing exploration and behavior diversity.
  \item \textbf{Implicit planning:} Diffusion steps act as an optimizer, guiding the model towards optimal behaviors conditioned on goals or rewards.
  \item \textbf{Compatibility with offline RL:} Since DMs learn from static datasets, they are well-suited for offline scenarios where online exploration is risky or expensive.
  \item \textbf{Constraint handling:} Classifier-guided sampling~\cite{dhariwal2021classifier} enables enforcing safety or energy constraints at inference time without modifying the base model.
\end{itemize}

\subsection{DMs for Multi-agent RL}\label{multi-agent-DRL}
DMs can also be extended for multi-agent and hierarchical learning. In multi-agent systems, RL becomes more complex due to non-stationary environments, coordination requirements, and communication constraints in POMDPs. DMs offer a new perspective by enabling the modeling of joint or decentralized policies with rich, expressive behavior distributions, which can improve cooperation, communication, and robustness, as shown in Table \ref{table-multi-agent-DRL-summary}. 

\subsubsection{Diffusion-based Joint Trajectory Generation}
\begin{figure}[!h]
    \centering
    \includegraphics[width=0.9\linewidth]{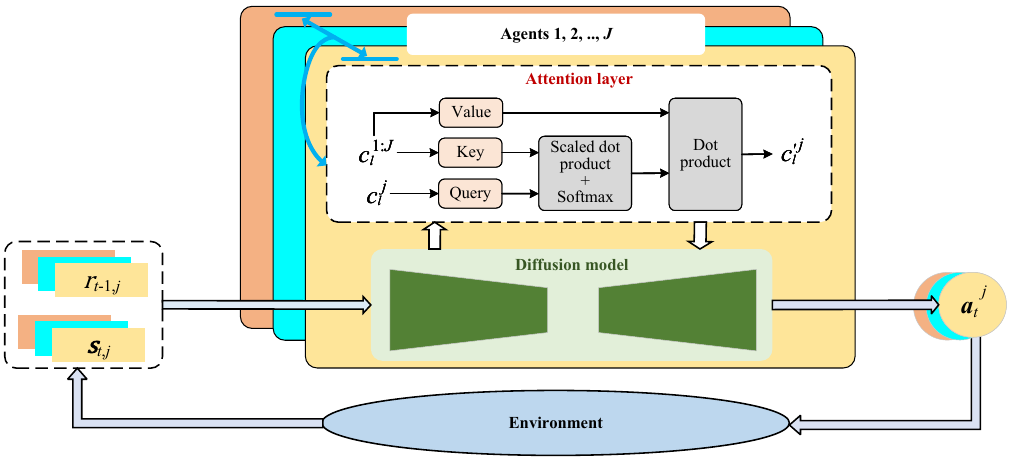}
    \caption{Framework of joint trajectory generation across agents with attention-based DMs. Every decoder layer of each agent performs attention across all agents. The multi-head attention mechanism is introduced to fuse the encoded feature $c'^{i}_{l}$ with other agents’ information for effective multi-agent collaboration.}
    \label{multi_agent_RL_Joint_Trajectory}
\end{figure}

For multi-agent systems and hierarchical tasks, they usually present additional challenges due to the need for coordinated planning and abstraction compared to single-agent and non-hierarchical learning \cite{du2021survey, pateria2021hierarchical}. DMs are well-suited for these settings due to their compositional nature. In centralized settings, the goal is to model the joint trajectory distribution across all agents \cite{jiang2023motiondiffuser}. Let $\tau = \{(s_1, \mathbf{a}_1), ..., (s_T, \mathbf{a}_T)\}$, where $\mathbf{a}_t = (a_t^1, ..., a_t^N)$ is the set of actions for $N$ agents. A DM is trained over the joint sequence to learn coordinated behavior:
\begin{equation}
    p_\theta(\tau^{(0)} | s_1, \mathbf{g}) = \prod\nolimits_{t=1}^{T} p_\theta(\mathbf{a}_t | \tau_{<t}, s_1, \mathbf{g}),
\end{equation}
where $\mathbf{g}$ may represent shared team goals. This approach enables trajectory-level coordination and planning, as seen in \textit{Multi-agent Learning Diffuser} \cite{zhu2024madiff}. The framework of joint trajectory generation across agents with attention-based DMs is shown in Fig. \ref{multi_agent_RL_Joint_Trajectory}. 

\subsubsection{Diffusion-based Collaborative Policy Learning}
\begin{figure*}[!h]
    \centering
    \includegraphics[width=0.9\linewidth]{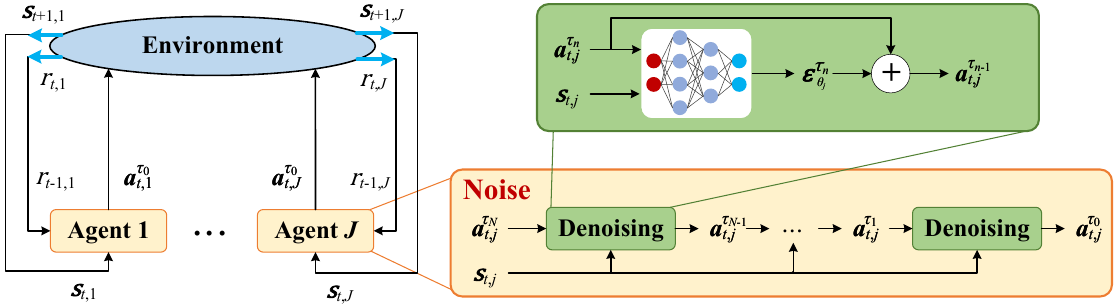}
    \caption{Framework of decentralized policy learning with diffusion-based multi-agent RL. Each agent generates the actions with a DM according to its observed system state.}
    \label{multi_agent_RL_Decentralized_Learning}
\end{figure*}

In multi-agent scenarios, each agent’s trajectory can be modeled as a conditional distribution over a shared latent space, allowing for collaborative policy learning \cite{yao2025enhancing, qi2024diffusion}. For instance, Li et al. \cite{li2023hierarchical} explore using diffusion to model sub-policy sequences, enabling flexible composition of high-level skills. These hierarchical DMs facilitate zero-shot generalization across tasks by generating coherent behavior segments that fulfill sub-goals. In decentralized settings, each agent independently learns a diffusion-based policy:
\begin{equation}
    a_t^i \sim p_\theta^i(a_t^i | o_t^i) = \text{ReverseDiffusion}_\theta^i(a_t^{i(T)} | o_t^i),
\end{equation}
where $o_t^i$ is agent $i$'s local observation. This allows agents to reason over multimodal action spaces, improving adaptability in competitive and partially observable environments. The framework of decentralized policy learning with diffusion-based multi-agent RL is shown in Fig. \ref{multi_agent_RL_Decentralized_Learning}.

\subsubsection{Theoretical and Practical Benefits}
These methods of the DMs for multi-agent RL have shown promise in domains such as StarCraft Multi-Agent Challenge (SMAC), multi-robot warehouse control, and autonomous driving platoons. Many practical benefits include:  
\begin{itemize}
  \item \textbf{Non-stationarity mitigation:} The conditional sampling process stabilizes policy updates across agents.
  \item \textbf{Better coordination:} Joint distribution modeling enhances synchronized behaviors in cooperative games.
  \item \textbf{Scalability:} Factorized sampling and latent conditioning reduce the burden of modeling large joint action spaces.
  \item \textbf{Offline multi-agent learning:} DMs can leverage offline datasets, such as team trajectories, to bootstrap decentralized policies.
\end{itemize}

\section{DMs for Online RL and Offline RL: Technique Taxonomy}\label{technique-taxonomy}
This section further discusses the DMs for RL from the perspective of technique taxonomy. Compared to the function taxonomy, the MDs play different roles and implementations in the technique taxonomy. In particular, we present the DMs for online RL and offline RL. Moreover, we discuss how to use the DMs to address the challenges of each category, along with theoretical perspectives.

\subsection{Diffusion-based Online RL}
Online RL refers to learning policies through continuous interaction with an environment, where agents collect new experiences and update their behavior in real time. While online RL is highly effective in adapting to dynamic environments, it often suffers from instability and sample inefficiency, particularly in settings with distributional shifts or sparse rewards. For example, in robotic locomotion tasks, a legged robot may encounter unmodeled terrain or sensor noise, which can lead to unstable updates if the policy overfits to recent experiences. Similarly, in autonomous driving, rare traffic scenarios such as sudden pedestrian crossings or vehicle malfunctions provide sparse feedback, making conventional online RL methods slow to adapt.

Recent works have introduced DMs to improve exploration, policy expressiveness, and trajectory diversity in online settings~\cite{luo2024text}. By modeling multimodal action distributions, DMs allow agents to consider multiple plausible futures, which can enhance both safety and performance. For example, in DiffusionQL~\cite{wang2023diffusion}, the Gaussian policy in Soft Actor-Critic (SAC) is replaced with a diffusion policy that captures complex, multimodal action distributions. This enables an agent navigating a maze to explore multiple paths simultaneously, improving the likelihood of discovering high-reward trajectories even under sparse feedback.

Similarly, in distributed systems, EdgeDiffusion \cite{xu2024accelerating} applies latent-action diffusion to online collaborative edge computing. Here, multiple edge servers must schedule tasks under uncertain workloads. The diffusion policy generates diverse candidate scheduling actions, allowing the system to balance load dynamically and improve throughput. This approach demonstrates how DMs can enhance online decision-making in non-robotic, high-dimensional action spaces where uncertainty is inherent.

Another key development is the use of consistency models to accelerate sampling. Song et al.~\cite{song2023consistency} propose a framework where the iterative denoising process is replaced with a single-step or few-step deterministic mapping, significantly reducing inference time. This is critical for high-frequency online decision-making, such as controlling a drone at 50–100 Hz or updating trading strategies in algorithmic finance, where standard diffusion sampling would be too slow. By combining rapid sampling with expressive multimodal distributions, consistency models allow diffusion-based policies to operate in fast-paced, real-world settings without sacrificing performance.

In addition to exploration and efficiency, DMs also offer advantages in robustness and generalization. For example, in autonomous driving simulations, a diffusion policy can sample multiple feasible trajectories when encountering unpredictable obstacles, reducing the chance of catastrophic failures. In robotic manipulation, DMs can propose alternative grasps or motion sequences when the initially planned trajectory is blocked or fails mid-execution.

In summary, the integration of DMs into online RL addresses several core challenges: enhancing exploration in sparse-reward environments, capturing multimodal action distributions, enabling efficient high-frequency inference, and improving robustness to uncertainty. These advances position diffusion-based policies as a promising approach for real-world online decision-making across robotics, distributed systems, autonomous driving, and other high-stakes applications.

\subsection{Diffusion-based Offline RL}
Offline RL focuses on learning policies from pre-collected datasets without any online interaction with the environment. While this setting enables safe and efficient policy learning, it introduces challenges such as distributional mismatch between the training dataset and the deployment environment, and extrapolation error, where the learned policy produces actions that lie outside the support of the offline data. For instance, a robot trained offline on a dataset of kitchen manipulation tasks may encounter novel object configurations or tools at deployment time, causing failures if the policy cannot generalize beyond the observed trajectories. Similarly, offline autonomous driving datasets may lack rare but critical events (e.g., sudden pedestrian crossings), making it difficult for the agent to handle such scenarios safely.

DMs have emerged as a powerful tool for offline RL by framing policy learning as trajectory modeling and conditional sequence generation. The Trajectory Diffuser~\cite{janner2022planning} is a foundational method that trains a DM on offline trajectories and performs trajectory planning by conditioning on target goals or rewards. For example, in MuJoCo locomotion tasks, the model can generate sequences of joint torques to achieve a specified endpoint or velocity, effectively planning multi-step behaviors without online interaction.

Building on this, Decision Diffuser~\cite{ajay2023conditional} treats offline decision-making as a conditional sequence generation problem, allowing agents to generate entire trajectories conditioned on desired outcomes. This approach has been applied to robotic manipulation, where the model can generate sequences of pick-and-place actions conditioned on the goal of arranging objects into a specific configuration, providing a flexible and general framework for complex offline tasks.

Motion Diffuser~\cite{zhang2024motiondiffuse} further extends diffusion-based planning to motion generation tasks, producing smooth and feasible motion trajectories for robots or animated characters from offline motion datasets. For instance, it can generate walking, running, or jumping sequences that respect physical constraints while interpolating between motion primitives observed in the training data.

Subsequent developments focus on improving generalization and control precision in offline settings. Trajectory-level sampling with classifier guidance\cite{ni2023metadiffuser, liang2023adaptdiffuser} allows the model to bias trajectory generation toward high-reward regions or satisfy safety constraints. For example, in warehouse robotics, classifier-guided sampling can steer motion plans away from obstacles while still achieving task goals efficiently. Diffusion-based latent skill priors\cite{kim2024robust, liang2024skilldiffuser} capture reusable behaviors as compressed latent representations, enabling the policy to compose skills flexibly. In multi-task robot manipulation, such priors allow the agent to adapt learned skills to novel objects or goals that were not present in the offline dataset.

Collectively, these methods demonstrate that offline diffusion-based RL can model complex, multi-step behaviors, handle high-dimensional action spaces, and incorporate goal or reward conditioning. They provide a foundation for safe and generalizable policy learning from static datasets, paving the way for applications in robotics, autonomous driving, and other real-world sequential decision-making tasks where online exploration is costly or unsafe.

\subsection{Theoretical Insights}
Beyond empirical performance, diffusion-based RL is being studied theoretically. Key questions include sample complexity, convergence guarantees, and variance analysis. Works like \cite{wang2023diffusion} and \cite{ajay2023conditional} analyze the stability and expressivity of diffusion-based policies compared to Gaussian baselines. Consistency training \cite{song2023consistency} provides provable guarantees for fast sampling. Further, connections between DMs and score-based RL have been explored to justify their decision-theoretic advantages \cite{sun2025score}. Ongoing research seeks to understand the limitations of denoising-based policies in high-dimensional or partially observable spaces \cite{kong2021fast, venkatraman2023reasoning, zhang2024offline}.

\section{Application Scenario in the DMs for RL}\label{sec-applications}
\begin{figure*}[!t]
    \centering
    \includegraphics[width=\linewidth]{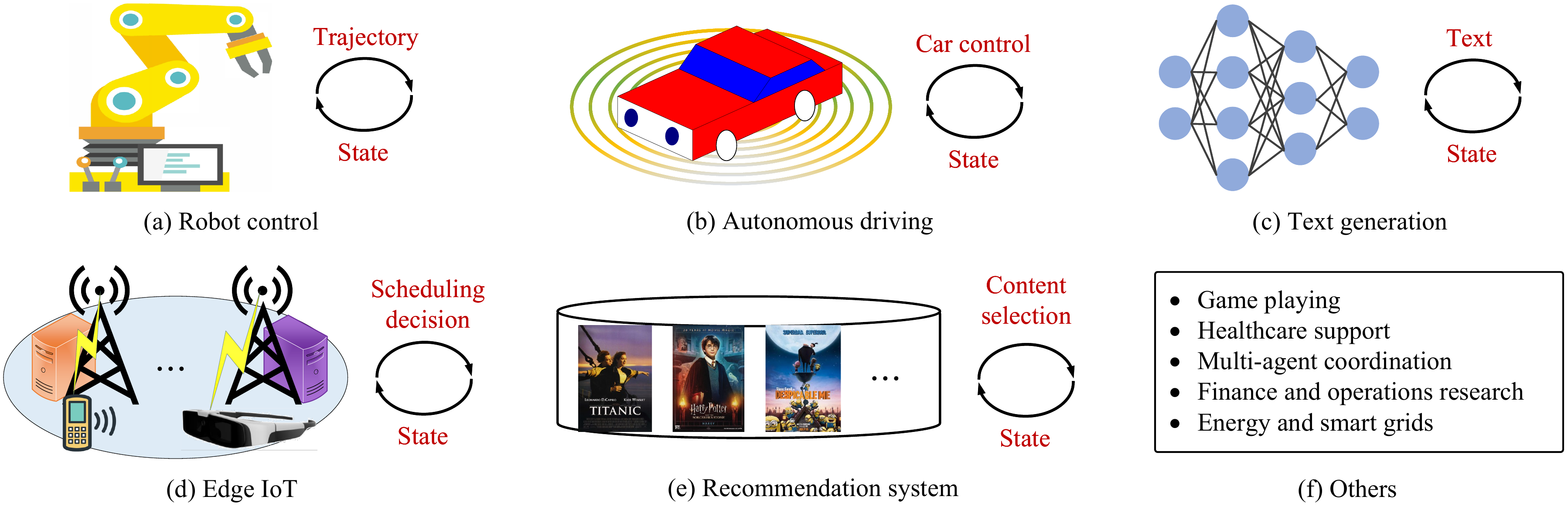}
    \caption{Illustration of application scenarios with diffusion-based RL.}
    \label{applications_figure}
\end{figure*}

The theoretical underpinnings of DMs in RL remain an active research area. One interpretation views diffusion sampling as a form of stochastic planning, where the denoising process incrementally corrects sampled sequences toward feasible and optimal solutions. The capacity of DMs to capture multimodal distributions, integrate prior knowledge, and generate diverse, high-quality samples makes them especially well-suited for real-world applications characterized by dynamic and uncertain environments. This section highlights recent progress in leveraging DMs across various application domains, mainly including robotic control \cite{liang2023adaptdiffuser}, autonomous driving \cite{liao2025diffusiondrive}, text generation \cite{zou2023survey}, edge Internet of Things (IoT) \cite{xu2024accelerating}, and recommendation systems \cite{lin2024survey}. An illustration of these application scenarios is presented in Fig.~\ref{applications_figure}.

\subsection{Robot Control}
Diffusion-based models have been increasingly applied to robotic control tasks, including locomotion, manipulation, and dexterous object handling \cite{wolf2025diffusion, chi2023diffusion, liu2025diffusion}. These tasks involve high-dimensional, continuous action spaces and require precise coordination over long time horizons, making them ideal candidates for generative modeling approaches. For example, the Diffuser framework introduced in \cite{janner2022planning} leverages a DM to generate action trajectories conditioned on both past and desired future observations. This enables goal-conditioned planning in continuous control environments, where traditional RL approaches often struggle with sparse rewards or suboptimal exploration.

Unlike conventional RL policies that directly map states to actions, DMs operate over the trajectory level, enabling a richer representation of behavior and facilitating the reuse of prior knowledge. Such models have demonstrated superior flexibility in capturing complex, multimodal action distributions and offer enhanced generalization to unseen states or goals. Moreover, they are particularly effective in scenarios where diverse skills must be combined to solve new tasks. For instance, \cite{ajay2023conditional} demonstrates that DMs trained on diverse behavior datasets can compose primitive behaviors in a zero-shot manner, enabling agents to adapt to novel tasks without requiring additional fine-tuning. This compositional ability is especially valuable in real-world robotics, where data collection is expensive and retraining policies from scratch is often impractical.

In addition, recent works have explored hybrid methods that integrate DMs with learning-based control frameworks, such as model predictive control, to achieve both long-term planning and real-time reactivity. These approaches demonstrate improved robustness to distributional shifts and physical perturbations, further highlighting the potential of diffusion-based RL models as a powerful paradigm for planning tasks in robotics.

\subsection{Autonomous Driving}
In autonomous driving applications, decision-making systems must reason under uncertainty, manage complex, multimodal interactions among agents, and consistently satisfy stringent safety requirements. These systems are required not only to predict a distribution over possible future trajectories but also to ensure that selected actions are both feasible and aligned with safety-critical constraints. Traditional methods such as rule-based planners or probabilistic models often fall short in capturing the full range of possible driving behaviors, particularly in highly interactive or uncertain environments \cite{deo2018convolutional, westny2023mtp}. 

To address these challenges, diffusion-based policies have been increasingly adopted for trajectory prediction and decision-making tasks in autonomous driving applications \cite{wang2025diffad, wang2024drivedreamer}. Owing to their inherent capacity to model complex, multi-modal distributions, DMs can generate a diverse set of plausible future trajectories that reflect the uncertainty and variability in driver intentions and surrounding traffic dynamics. This capability is especially valuable in tasks such as trajectory forecasting, where anticipating multiple likely outcomes enables more robust downstream planning. Beyond unconditional generation, recent research has explored the integration of control constraints directly into the sampling process using classifier-guided diffusion. These methods leverage auxiliary classifiers to bias the denoising steps toward safe and legal outcomes, effectively embedding constraints such as collision avoidance, lane-keeping, and traffic rule compliance into the trajectory generation process \cite{brehmer2023edgi}. This integration allows DMs to maintain flexibility in behavior generation while ensuring adherence to operational safety constraints. Moreover, the ability of DMs to operate over sequences of future actions rather than individual decisions enables holistic planning that accounts for long-term dependencies—an essential feature in complex traffic scenarios involving merges, roundabouts, or pedestrian interactions. The stochastic nature of the sampling process also facilitates risk-aware planning by providing access to an ensemble of trajectories that can be used to evaluate uncertainty and worst-case outcomes.

As autonomous driving applications continue to evolve toward higher levels of autonomy and urban deployment, diffusion-based RL techniques are poised to play a critical role in improving the robustness, flexibility, and safety of real autonomous driving systems.
 
\subsection{Text Generation}
In text generation scenarios, DMs have recently been explored as powerful alternatives to traditional autoregressive language models for tasks involving sequential generation, such as dialogue generation, machine translation, and story completion \cite{he2022diffusionbert, nachmani2021zero, gong2023diffuseq}. Unlike existing autoregressive models, which generate text token-by-token in a left-to-right fashion and are prone to exposure bias and compounding errors, diffusion-based NLP models enable a non-autoregressive framework where entire sequences are refined iteratively through denoising steps. This process allows for greater global control over the generated content and reduces error propagation.

DMs excel in capturing global dependencies and long-range semantic coherence across a sequence, which is particularly advantageous in applications like multi-turn dialogue and story generation \cite{zou2023survey}. In these contexts, it is crucial for the model to maintain contextual consistency, handle ambiguous user inputs, and align generated responses with the overarching conversation intent. DMs can be conditioned on rich contextual embeddings or future constraints, allowing them to generate responses that are not only fluent but also semantically aligned with specified goals or outcomes \cite{li2022diffusionlm}. Additionally, diffusion-based language models have shown improved controllability, making them well-suited for constrained generation tasks where fine-grained control over syntax, semantics, or structure is required \cite{zhu2023diffusionnlp}. For instance, in story generation, users may wish to enforce narrative elements such as specific plot points, character development, or emotional tone. DMs allow these constraints to be incorporated during the generation process, either through conditional embeddings or classifier guidance, without requiring extensive retraining. Moreover, these models have demonstrated robustness in handling diverse input formats and can generalize well in few-shot or low-resource settings due to their ability to model uncertainty and sample diverse outputs. This diversity is particularly valuable in dialogue systems, where multiple valid responses may exist, and generating varied yet relevant replies is key to user engagement.

Overall, by combining expressive generative capacity with flexible conditioning and control mechanisms, DMs offer a promising paradigm for future advancements in text generation, especially in tasks that demand coherent long-form text generation, intent alignment, and adaptive decision-making over sequences.

\subsection{Edge IoT}
Edge computing has emerged as a promising paradigm for enabling low-latency services for computation-intensive IoT applications such as smart surveillance and real-time industrial control \cite{xu2024dynamic, liang2025collaborative, zou2024fine}. By bringing computational resources closer to data sources at the network edge, edge computing significantly reduces response time and alleviates the burden on centralized cloud infrastructure. However, the dynamic and heterogeneous nature of edge environments, characterized by varying network conditions, resource constraints, and task arrival patterns, poses substantial challenges to efficient task scheduling, service placement, and resource management.

Conventional approaches to edge task scheduling typically involve heuristic optimization or RL-based methods. While heuristics are computationally efficient, they rely on rigid, handcrafted assumptions about system states and fail to generalize well in dynamic environments \cite{xu2024incorporating}. RL methods, though more adaptive, often suffer from sample inefficiency, limited exploration in high-dimensional action spaces, and training instability due to non-stationary objectives and reward signals \cite{zhu2023diffusionrl}. These limitations hinder their ability to handle the complex coordination and decision-making required in multi-node edge computing scenarios. To overcome these bottlenecks, diffusion-based models have recently been explored for task scheduling in edge intelligence systems, offering a more expressive and stable decision-making framework \cite{du2024diffusion, zhang2025diffusion}. These models leverage the generative capabilities of diffusion processes to capture the distributional patterns of optimal scheduling trajectories, enabling diverse exploration and improved generalization. 

For instance, Xu \textit{et al.} \cite{xu2024accelerating} propose a latent action diffusion framework integrated into a deep RL model for task scheduling in collaborative edge computing environments. This method introduces a denoising diffusion process in the latent action space, which enhances the exploration of potential scheduling decisions while maintaining temporal coherence across decision steps. By operating in the latent space, the model effectively avoids the pitfalls of modeling raw action distributions directly, resulting in more robust and stable policy behavior. The framework demonstrates significant improvements in minimizing service delays and task processing latency across distributed edge nodes. Moreover, the proposed method’s ability to reduce policy instability and performance variance makes it especially well-suited for real-world edge-edge collaboration scenarios, where system dynamics and workloads can vary drastically over time. As edge computing infrastructures grow more decentralized and task offloading scenarios become increasingly complex, diffusion-based scheduling frameworks are poised to play a crucial role in achieving scalable, efficient, and adaptive service provisioning.

\subsection{Recommendation Systems}
Recommendation systems can benefit significantly from the application of DMs, particularly in scenarios that involve modeling user behavior sequences over time \cite{wang2023diffusionRM, ma2024multimodal, yang2023generate, wang2024leadrec}. Traditional recommendation methods—such as matrix factorization, collaborative filtering, and sequence-based deep learning models—often struggle to capture the stochasticity, multi-modality, and temporal dependencies inherent in user interaction patterns. In contrast, DMs offer a flexible generative framework capable of learning complex sequential distributions, making them particularly well-suited for synthesizing diverse recommendation trajectories that balance both short-term user preferences and long-term engagement objectives.

By treating user interaction histories as sequences of states and modeling their evolution through a denoising process, DMs enable the generation of future recommendation paths that are not only diverse but also coherent with past behavior. This capacity is especially useful in dynamic environments like e-commerce, streaming platforms, and personalized education, where user preferences shift over time and require adaptable, forward-looking strategies. Moreover, retrieval-augmented diffusion frameworks \cite{blattmann2022retrieval} push the boundaries of personalization by integrating external memory modules or retrieval mechanisms into the generative process. These systems retrieve semantically relevant historical sequences—either from the same user or from similar user profiles—and use them as additional conditioning signals during sampling. This hybrid approach enhances the diversity and relevance of generated recommendations by grounding them in real user behavior patterns while still allowing the flexibility of generative modeling.

Such frameworks also enable zero-shot or few-shot recommendation capabilities, as the retrieval component provides contextual grounding even in sparsely populated user-item interaction spaces. Overall, the integration of DMs into recommendation systems promises not only improved predictive performance but also a deeper understanding of sequential user behavior, paving the way for more robust, personalized, and context-aware recommendation engines.

\subsection{Others}
\textbf{Game Playing:} DMs have also been applied in game environments where the agent must learn to act based on sparse rewards and delayed feedback \cite{valevski2024diffusion}. The generative nature of DMs allows for modeling complex strategies and exploring diverse behaviors beyond the reach of conventional policies. For example, in open-world simulation games, diffusion policies have shown promise in generating long-term strategic behaviors that align with high-level goals \cite{alonso2024diffusion}.
\textbf{Healthcare Support:} Healthcare decision-making often involves long-term planning under partial observability and uncertainty. Diffusion-based sequential models can be used to simulate treatment trajectories or recommend interventions by learning from expert demonstrations. Their ability to generate plausible yet diverse treatment sequences aids in discovering alternative care plans, which is valuable in personalized medicine \cite{kazerouni2023diffusion, al2024diffusion}.
\textbf{Multi-Agent Coordination:} DMs have also been applied to cooperative and competitive multi-agent settings, where generating coordinated strategies is challenging. Recent works leverage the ability of DMs to jointly sample trajectories of multiple agents while maintaining temporal and spatial consistency \cite{zhu2024madiff}. This is particularly useful in applications such as swarm robotics and multiplayer gaming environments \cite{geng2023diffusion}.
\textbf{Finance and Operations Research:} In finance, sequential decision-making is crucial for tasks such as portfolio optimization and algorithmic trading. DMs can be used to simulate market trajectories and optimize trading strategies under uncertain conditions. Their ability to represent complex, multimodal distributions makes them suitable for capturing the stochastic nature of financial time series \cite{sattarov2023findiff}.
\textbf{Energy and Smart Grids:} In smart grid systems, energy management decisions must consider fluctuating demands and supply conditions over time. Diffusion-based policies can learn robust scheduling strategies that adapt to uncertainties, reduce energy consumption, and stabilize grid operations. These models are also applicable in load balancing and demand response scenarios \cite{lin2024energydiff, geng2025diffusion}.

\section{Open Research Issues and Future Directions} \label{open-issues-directions}
Despite the promising advances of DMs for RL, several critical challenges and open research issues remain. In this section, we outline these issues and highlight potential directions for future exploration, summarized in Table \ref{table-open-isusses-future-work-summary}.

\begin{table*}[!t]
\caption{Summary of representative papers on open research issues and future directions.}
\centering
\renewcommand{\arraystretch}{1.5}
\belowrulesep=0pt
\aboverulesep=0pt
\footnotesize 
\begin{tabular}{m{50pt}<{}|m{130pt}<{}|m{140pt}<{}|m{30pt}<{}}
\toprule
\textbf{Open Issues} & \textbf{Main Feature} & \textbf{Future Directions} & \textbf{Ref.}\\
\midrule
    Sampling efficiency & DMs have a high computational cost to adoption in real-time or high-throughput RL applications. &  Algorithmic innovation and systems-level optimization to enable practical deployment in time-sensitive applications. & \cite{midjourney, huggingface, song2021denoising, lu2022dpm}.\\
    \midrule
    Sampling variance & The stochastic nature of diffusion sampling introduces variability. & Explore variance-reduced or guided sampling strategies to enhance reliability and stability& \cite{haarnoja2018soft, song2023consistency, dhariwal2021diffusion} \\
    \midrule
    Hardware adaptation and energy efficiency & DMs typically impose high computational and memory demands. & Develop energy-aware and hardware-adaptive DMs across compression, quantization, and architecture design. & \cite{lyu2022accelerating}.\\
    \midrule
    Safety and ethical constraints & DMs rely heavily on post-hoc safety corrections. & Integrate operational constraints
and risk measures into DMs' generative process. & \cite{jiang2023motiondiffuser}.\\
    \midrule 
     Partial observability and uncertainty & DMs often assume full observability, sampling trajectories directly from state-action distributions. & Integrate belief states into the denoising process and employ memory-augmented architectures. & \cite{igl2018deep}.\\
    \midrule
    Long horizons and sparse rewards & DMs often accumulate compound errors in long sequences due to the iterative denoising process. & Explore hierarchical modeling and trajectory compression for DMs. & \cite{xu2024beyond} . \\
    \midrule
    Theoretical foundations and guarantees & The theoretical understanding of diffusion-based RL models remains limited. & Tackle the questions of expressivity, convergence, and generalization for diffusion-based RL. & \cite{xu2022prompting}.\\
    \midrule
    Benchmarks and standardized evaluation & Current methods rely on customized experimental setups, making it difficult to compare results across different tasks. & Establish the standardized benchmarks to cover a wider range of environments and tasks. & \cite{janner2022planning, liao2025diffusiondrive, gulcehre2020rl}.\\
    \midrule
    Online and continual learning & Most diffusion-based RL methods are tailored for offline settings with fixed datasets. & Explore efficient adaptation mechanisms, exploration-exploitation trade-offs, and robust noise scheduling strategies. & \cite{zhou2024redi, blattmann2022retrieval} .\\
   \midrule
    Multi-agent and human-in-the-loop systems & The complexity of multi-agent and human-interactive environments introduces unique challenges for DMs. & Blend diffusion-based generation with communication protocols, intention modeling, and coordination mechanisms. & \cite{meng2023offline}.\\
    \midrule
    Large language models & The integration of LLM and DM represents a promising frontier in various domains. & Enable DMs to handle complex inputs and generate multiple outputs with better performance in LLM scenarios. & \cite{li2022diffusionlm, wang2024drivedreamer, zhu2023diffusionnlp, wei2025automated}.\\
\bottomrule
\end{tabular}
\label{table-open-isusses-future-work-summary}
\end{table*}

\subsection{Improving Sampling Efficiency}
One of the primary limitations of DMs is their high computational cost, which stems from the iterative denoising process required for sample generation. For example, on commercial platforms such as Midjourney \cite{midjourney} and Hugging Face \cite{huggingface}, users often experience delays of $40$–$80$ seconds for producing a single image, even with access to powerful GPUs. This latency arises because traditional DDPMs typically rely on hundreds to thousands of reverse diffusion steps, each involving a forward pass through a large neural network. In practice, the network architectures are also massive (e.g., Stable Diffusion 3 contains on the order of $8$ billion parameters), which further exacerbates memory consumption and inference time.

Such computational demands are particularly problematic in RL, where policies must generate actions at millisecond timescales to interact with dynamic environments. In robotics, for example, a robot arm performing real-time grasping cannot afford delays of several seconds between action samples. Similarly, in autonomous driving, where control loops typically run at $10$–$100$ Hz, the requirement of hundreds of denoising iterations per action renders vanilla DDPMs impractical. High-frequency domains such as stock trading agents or multi-agent swarm control would also become infeasible if each decision step incurred multi-second delays. Although acceleration methods such as DDIM \cite{song2021denoising} and DPM-Solver \cite{lu2022dpm} significantly reduce the number of required denoising steps (e.g., from $1000$ to as few as $10$–$20$), challenges remain in balancing speed and sample quality. For instance, fewer steps can lead to degraded trajectory fidelity in decision-making tasks, which may translate into suboptimal or unsafe actions in RL settings. As a result, future research must prioritize the development of lightweight and efficient diffusion variants that support fast inference while retaining robustness and accuracy. Promising directions include model distillation techniques that compress iterative samplers into single-step or few-step generators, hybrid architectures that combine diffusion backbones with autoregressive or flow-based components, and hardware-aware optimizations leveraging edge accelerators or quantization.

In summary, while DMs have demonstrated remarkable generative capabilities, their computational overhead remains a significant barrier to adoption in real-time or high-throughput RL applications. Bridging this gap requires both algorithmic innovation and systems-level optimization to enable practical deployment in time-sensitive RL domains.

\subsection{Reducing Sampling Variance}
The stochastic nature of diffusion sampling introduces variability that can hinder performance in tasks requiring deterministic or low-variance behaviors, such as robotics and safety-critical control. For instance, in robotic manipulation tasks, small variations in the sampled trajectory can result in unstable grasps or collisions with the environment, while in autonomous driving, stochastic deviations in steering or braking actions may compromise safety margins. Unlike conventional RL policies that default to the mean of a Gaussian distribution for stable action selection \cite{haarnoja2018soft}, DMs inherently depend on random initialization (sampling from Gaussian noise) and a sequence of stochastic denoising updates, leading to non-negligible trajectory variance across runs.

This variability poses challenges when high precision and repeatability are required. For example, in industrial assembly robots, repeating the same motion plan should yield nearly identical executions, whereas a diffusion-based policy may introduce subtle but critical differences due to randomness in its sampling path. Similarly, in safety-critical domains like UAV navigation in cluttered environments, even minor stochastic deviations in planned trajectories can cause collisions with obstacles.

Future work should therefore explore variance-reduced or guided sampling strategies to enhance reliability and stability. Promising directions include consistency models \cite{song2023consistency}, which eliminate the need for iterative stochastic denoising and instead learn a deterministic mapping from noise to sample in a single or few steps, thereby reducing randomness in the generation process. Another direction is classifier-guided sampling \cite{dhariwal2021diffusion}, where external guidance functions (e.g., reward gradients, safety classifiers, or control constraints) are incorporated during sampling to bias trajectories toward more reliable or constraint-satisfying behaviors. Beyond these, domain-specific strategies such as incorporating trajectory smoothing, using deterministic initialization seeds, or hybridizing diffusion outputs with conventional low-variance RL policies may further improve robustness.

\subsection{Hardware-Aware and Energy-Efficient Design}
As is well known, DMs typically impose high computational and memory demands, which severely hinder their deployment on resource-constrained systems, such as mobile robots, unmanned aerial vehicles, or edge devices. For example, even state-of-the-art latent DMs often require several gigabytes of GPU memory and tens of billions of floating-point operations per sampling trajectory. Such requirements exceed the capabilities of lightweight hardware platforms like NVIDIA Jetson boards, which are commonly used in mobile robotics or edge AI accelerators in IoT devices. In these settings, high energy consumption not only limits inference throughput but also drains battery life, posing practical challenges for long-duration autonomy.

This issue is particularly acute in domains requiring on-device decision-making. For instance, an autonomous drone tasked with obstacle avoidance must infer collision-free trajectories within milliseconds on its onboard processor, where GPU resources are minimal. Similarly, service robots deployed in households or warehouses must run policies locally to ensure reliability and privacy, but the large memory footprint and latency of DMs make this infeasible without optimization. In edge computing for smart transportation, deploying DMs for vehicle trajectory prediction or traffic flow control is constrained by limited bandwidth and the need for energy-efficient inference across many distributed nodes.

To address these limitations, future research should further investigate model compression strategies, such as pruning redundant diffusion layers or distilling iterative sampling into lightweight one-step generators~\cite{lyu2022accelerating}. Quantization techniques, which reduce the precision of parameters (e.g., from FP32 to INT8), can substantially lower memory usage and power consumption while retaining accuracy, making DMs more compatible with specialized hardware accelerators. Additionally, architectural optimizations, for example, designing smaller U-Net backbones, leveraging lightweight attention modules, or adopting mobile-friendly operators such as depthwise separable convolutions, could reduce both energy consumption and inference latency. Promising directions also include edge-cloud collaborative frameworks, where expensive denoising steps are offloaded to the cloud while lightweight local inference handles fast decision-making.

In summary, enabling DMs to operate efficiently on mobile and edge platforms requires innovations across compression, quantization, and architecture design. Developing energy-aware and hardware-adaptive DMs will be critical for unlocking their potential in real-world decision-making applications, particularly in robotics, autonomous systems, and large-scale IoT environments.

\subsection{Integration with Safety and Ethical Constraints}
Safe deployment of AI agents requires not only high performance but also strict compliance with operational and safety constraints. For example, in autonomous driving, a planning agent must ensure collision avoidance and adherence to traffic rules, while in healthcare, treatment recommendation systems must avoid unsafe dosage levels or harmful interventions. Similarly, in robotic manipulation, agents must respect torque and force limits to prevent damage to both the robot and its environment. Violating such constraints can lead to catastrophic outcomes, highlighting the need for diffusion-based decision-making models that are inherently safety-aware.

While classifier guidance has shown promise for post-hoc constraint incorporation by steering generated samples toward safe regions during inference \cite{dhariwal2021diffusion}, this approach remains limited in real-time or safety-critical scenarios. For instance, in robot motion planning, classifier-guided diffusion may correct unsafe trajectories, but it cannot guarantee that all intermediate rollouts respect joint or velocity constraints, which could be unacceptable in physical systems.

To address these limitations, future work should explore integrating safety-aware objectives directly into the training process~\cite{wang2024safe}. For example, reward shaping could penalize safety violations during denoising, encouraging the DM to learn distributions that inherently avoid unsafe actions. Similarly, constraints such as maximum acceleration or energy budgets could be encoded as additional conditioning signals, ensuring the safety of the denoising trajectory.

Adapting frameworks such as constrained MDPs or risk-sensitive RL to the diffusion setting is another promising direction. In constrained MDPs, constraints like collision probability thresholds can be explicitly incorporated into the optimization objective. Extending this framework to diffusion-based agents would allow trajectory sampling to be conditioned on both reward maximization and constraint satisfaction. Likewise, risk-sensitive RL methods, which account for uncertainty by optimizing for conditional value-at-risk or other safety metrics, could be adapted to DMs to ensure robust decision-making under stochasticity. For example, in financial trading applications, a diffusion-based policy could be trained to generate trading strategies that not only maximize profit but also limit downside risk exposure.

In summary, while current diffusion approaches rely heavily on post-hoc safety corrections, future research should move toward intrinsically safe DMs that integrate operational constraints and risk measures directly into their generative process. This shift is essential for deploying diffusion-based agents in domains such as autonomous vehicles, robotics, healthcare, and finance, where safety is not optional but mandatory.

\subsection{Handling Partial Observability and Uncertainty}
Real-world tasks in RL often involve uncertainty and partial observability, where agents do not have access to the full environment state. Such scenarios are typically modeled as POMDPs, which are pervasive in applications like robotics, healthcare, and autonomous navigation. For instance, a mobile robot navigating through a smoke-filled environment may only perceive local LiDAR or camera readings, while critical parts of the global map remain hidden. Similarly, in healthcare, a treatment policy must be derived from incomplete and noisy patient records, where not all physiological variables are observable. In autonomous driving, occlusions (e.g., vehicles hidden behind buildings or pedestrians crossing from blind spots) exemplify the challenges of RL under partial observability.

Extending DMs to POMDPs is therefore essential for enhancing their applicability in such real-world scenarios. Current DM-based approaches often assume full observability, sampling trajectories directly from state-action distributions. However, in POMDPs, the agent must maintain and update a belief state (a probability distribution over possible hidden states) to make robust decisions. Without this capability, diffusion-based policies may generate infeasible or unsafe trajectories when crucial information is missing.

Future directions could include the integration of belief states into the denoising process. For example, a DM could be conditioned not only on observed trajectories but also on a compact representation of the agent’s belief about the hidden environment. This would allow the generative process to account for uncertainty explicitly, leading to safer and more robust decisions. Alternatively, memory-augmented architectures could be employed, where recurrent modules (e.g., LSTMs or GRUs) or attention-based mechanisms (e.g., transformers) are combined with DMs to capture temporal dependencies and encode histories of past observations. Such hybrid approaches could help the model infer latent dynamics even when current observations are insufficient.

These directions are inspired by prior work in POMDP policy learning, such as deep recurrent Q-networks and transformer-based policies~\cite{igl2018deep}, which demonstrated that augmenting policies with memory significantly improves performance in partially observable settings. For example, in multi-agent pursuit-evasion games, a diffusion-based planner enhanced with a transformer could maintain a belief about the evader’s hidden location, enabling more coordinated strategies among pursuers. Similarly, in dialogue systems, where only partial user intent is observable, a memory-augmented DM could generate more coherent and contextually appropriate responses by leveraging conversation history.

In summary, bridging DMs with POMDP frameworks through belief modeling, recurrent memory, or attention-based mechanisms represents a crucial step toward deploying diffusion-driven RL in real-world, uncertainty-rich environments.

\subsection{Scaling to Long Horizons and Sparse Rewards}
Long-horizon planning and sparse rewards present major challenges in RL, primarily due to the difficulty of credit assignment—determining which past actions are responsible for delayed outcomes. For example, in a robot navigation task, the reward may only be received upon reaching a distant goal after hundreds of time steps. Similarly, in strategy games such as StarCraft II, winning or losing is determined by thousands of intermediate decisions, making it difficult to attribute credit to specific actions.

In such scenarios, DMs can struggle, as the iterative denoising process tends to accumulate compounding errors across long sequences. A small deviation early in the trajectory may cascade into significant divergence from the optimal policy. For instance, if a DM generates slightly suboptimal steering actions in the first few steps of an autonomous driving task, the vehicle may drift toward unsafe areas, and these errors can amplify over the course of the trajectory.

To address these challenges, hierarchical approaches have been explored. For example, DiffusionDrive \cite{liao2025diffusiondrive} decomposes planning into a high-level DM that generates subgoals (e.g., waypoints) and a low-level controller that executes primitive actions toward these subgoals. This hierarchical abstraction reduces the effective planning horizon for each model, making credit assignment more tractable. Similarly, efficient hierarchical diffusion methods \cite{kang2023efficient} structure long-horizon decision-making by first generating a coarse-grained trajectory outline, then refining it into fine-grained actions. Such approaches not only improve performance in navigation and locomotion tasks but also enhance interpretability, as subgoals provide a natural explanation for the model’s decisions.

Another promising direction is trajectory compression techniques, where long sequences are mapped into compact latent representations. For example, instead of generating hundreds of fine-grained robot joint movements, a DM could generate a compressed sequence of motion primitives (e.g., “move forward,” “turn left,” “pick up object”) that are then decoded into detailed motor commands. This approach is analogous to skill discovery in reinforcement learning, where reusable behaviors (skills) are learned and recombined to solve complex tasks. In the context of sparse-reward tasks like robotic manipulation, trajectory compression can help the DM focus on high-level progress (e.g., reaching and grasping) rather than noisy low-level dynamics.

As a concrete example, in autonomous warehouse robots, a DM could hierarchically plan routes across multiple aisles (high-level path planning) and then generate detailed wheel control signals (low-level execution). By abstracting the problem into levels of granularity, the DM avoids compounding small errors across hundreds of time steps and improves robustness in real-world deployments. Similarly, in multi-step medical treatment planning, a DM could generate a sequence of intermediate treatment milestones (e.g., stabilize vital signs, reduce infection) before filling in the finer details of dosage and scheduling.

In summary, tackling long-horizon credit assignment in DMs requires strategies such as hierarchical modeling and trajectory compression, which reduce complexity by structuring the output space into manageable skills or abstractions, thus improving scalability and robustness for RL.

\subsection{Theoretical Foundations and Guarantees}
Despite their empirical success, the theoretical understanding of diffusion-based RL models remains limited. Fundamental questions about expressivity, convergence, and generalization are still largely unanswered. Unlike conventional policy gradient or value-based RL methods, where convergence properties can often be established under certain assumptions (e.g., convexity or Lipschitz continuity), DMs involve iterative denoising with high-dimensional stochastic dynamics, making it difficult to derive formal guarantees.

One open question concerns the expressivity of DMs as policy classes. While DMs have shown strong capabilities in modeling complex trajectory distributions, it is still unclear what classes of policies or trajectory distributions can be efficiently represented. For example, in robotic manipulation, DMs can generate smooth trajectories that capture multimodal strategies (e.g., grasping an object from different angles). However, whether DMs can universally approximate any trajectory distribution, or under what assumptions they outperform Gaussian mixture models or autoregressive policies, remains an open theoretical problem.

Convergence analysis is another challenge. Standard RL policies typically update parameters toward fixed points defined by Bellman operators, which offers at least a pathway to convergence guarantees. In contrast, diffusion-based RL involves sampling-based optimization over a denoising process with dozens or hundreds of stochastic steps. This raises questions about whether training is guaranteed to converge to an optimal policy, or if it risks collapsing to degenerate solutions. For instance, in offline RL, where policies are trained from logged datasets, DMs might overfit to high-density regions of the data while ignoring rare but critical transitions. Understanding whether training dynamics can avoid such pitfalls is a crucial direction.

A further open issue lies in generalization. In real-world deployment, agents often face environments with significant distribution shifts. For example, an autonomous drone trained in a controlled simulation might need to generalize to outdoor environments with wind, sensor noise, and moving obstacles. While DMs excel at modeling training distributions, it is unclear whether the learned denoising dynamics can generalize effectively to unseen domains or whether they will degrade due to compounding errors across timesteps. Existing work such as prompting-based analysis~\cite{xu2022prompting} provides early insights into how conditional inputs guide diffusion sampling, but does not yet address generalization bounds in sequential decision-making.

Finally, the lack of a rigorous theoretical foundation also limits interpretability and safety analysis. In safety-critical domains, such as autonomous driving or healthcare decision-making, it is essential to provide guarantees about worst-case performance, stability, or risk sensitivity. Without a theoretical basis, practitioners must rely on empirical testing, which may fail to capture rare but catastrophic failure modes. For example, a diffusion policy that works well in 99\% of driving scenarios could still fail unpredictably in rare edge cases (e.g., unusual pedestrian behavior), undermining trust and deployment readiness.

In summary, while diffusion-based RL models demonstrate impressive empirical results, the field lacks a solid theoretical grounding. Future research must tackle questions of expressivity (what policies DMs can represent), convergence (whether learning dynamics are stable), and generalization (how models behave under distribution shifts). Addressing these challenges will not only improve interpretability and safety but also build confidence for deploying DMs in real-world sequential decision-making tasks.

\subsection{Benchmarks and Standardized Evaluation}
There is a significant lack of standardized benchmarks to evaluate diffusion-based RL methods. Most current studies rely on customized experimental setups, making it difficult to compare results across different works. For example, some papers test on simple locomotion tasks such as HalfCheetah-v2 or Hopper-v2, while others focus on offline robotic datasets (e.g., Franka kitchen manipulation). Without a consistent evaluation protocol, it is challenging to assess whether improvements stem from the model itself or from differences in datasets, reward shaping, or implementation details.

Community efforts toward standardization have begun but remain limited. For instance, Diffuser\cite{janner2022planning} introduced a framework for trajectory optimization in decision-making using DMs, primarily evaluated on MuJoCo locomotion tasks. Similarly, Decision Diffuser \cite{liao2025diffusiondrive} extended this line to autonomous driving scenarios, offering insights into multi-agent interactions and long-horizon planning. On the reinforcement learning side, large-scale offline RL datasets such as RLUnplugged~\cite{gulcehre2020rl} provide a diverse set of benchmarks for policy evaluation, but they are not specifically tailored to diffusion-based methods and often lack tasks that stress long-horizon credit assignment, partial observability, or safety-critical decision-making.

Concrete examples highlight this gap. In robotics, diffusion policies are often tested in simulated environments like D4RL’s kitchen task, where agents must open a microwave or move a kettle. However, these settings are relatively constrained compared to real-world robot deployments involving unstructured environments, human interaction, and safety constraints. In autonomous driving, datasets like Waymo Open Motion or nuScenes contain rich trajectory information that could serve as a benchmark, but diffusion-based decision-making studies have only selectively adopted them, often preprocessing or downsampling to fit their specific training pipeline. Similarly, in multi-agent RL, benchmarks such as StarCraft II micromanagement or Overcooked-AI are widely used in RL but have not yet been systematically adapted to diffusion-based approaches.

Therefore, future work should expand benchmark initiatives to cover a wider range of environments and tasks as follows. \textit{1) Robotics:} Standardized manipulation and navigation benchmarks with both simulation and real-world evaluation. \textit{2) Autonomous Driving:} Trajectory forecasting and decision-making benchmarks with safety metrics. \textit{3) Multi-Agent Systems:} Cooperative and competitive benchmarks where DMs must handle coordination and communication. \textit{4) Safety-Critical Domains:} Healthcare decision-making or industrial control, where robustness and constraint satisfaction are essential.

Establishing such standardized benchmarks would not only enable fair comparison across methods but also accelerate progress by identifying the strengths and limitations of diffusion-based RL in diverse application domains.

\subsection{Extending to Online and Continual Learning}
Most diffusion-based methods are tailored for offline settings with fixed datasets, where the model learns from large, static collections of trajectories (e.g., D4RL benchmarks for locomotion or manipulation). While effective in these controlled contexts, such approaches face significant challenges in real-world scenarios, where agents must adapt continuously to non-stationary data distributions caused by changing environments, dynamics, or task objectives. For example, a household robot trained offline on kitchen manipulation may fail if new objects, layouts, or user preferences are introduced. Similarly, in autonomous driving, policies trained on historical datasets may degrade when traffic rules change or new driving patterns emerge.

Adapting DMs to the dynamic nature of online RL remains an open problem~\cite{zhou2024redi}. Unlike offline settings where the distribution is fixed, online RL requires constant data collection, exploration, and adaptation. This creates challenges for DMs since their iterative denoising process is computationally intensive, making rapid policy updates difficult. Moreover, online interaction exacerbates the risk of compounding errors if the model fails to generalize outside the training distribution.

To address these challenges, meta-learning and retrieval-augmented strategies show promise. For example, meta-learning could allow DMs to learn adaptation priors, enabling rapid fine-tuning when faced with novel tasks, much like how Model-Agnostic Meta-Learning (MAML) accelerates adaptation in conventional RL. Retrieval-augmented approaches~\cite{blattmann2022retrieval}, where relevant past experiences are dynamically queried to guide sampling, could help diffusion policies remain robust under distribution shifts. For instance, in robotic manipulation, retrieval-based DMs could reference trajectories involving similar object configurations to stabilize learning in unfamiliar scenarios.

At a more conceptual level, the iterative refinement inherent to the denoising process can be interpreted as performing gradient-like updates in a latent trajectory space. This analogy provides a new perspective on convergence and expressiveness, suggesting that DMs are implicitly optimizing trajectories toward high-likelihood or high-reward regions of the behavior space. However, this interpretation raises important open questions. For example: \textit{1) Sample Complexity:} How many online interactions are needed for DMs to achieve stable convergence compared to policy gradient or Q-learning methods? \textit{2) Stability:} Can denoising dynamics remain stable in online settings where the trajectory distribution shifts rapidly? \textit{3)
Noise Scheduling:} How does the choice of forward diffusion noise schedule influence the balance between exploration and exploitation in online environments? 

Recent efforts are beginning to explore these issues. For instance, researchers have experimented with hybrid architectures that integrate diffusion-based policies with value functions to stabilize training and guide exploration, similar to how actor-critic methods balance policy learning and value estimation. In robotics, this could mean using a diffusion policy to propose diverse candidate trajectories while a value function evaluates their feasibility and reward. In autonomous driving, online adaptation could involve diffusion policies generating trajectory candidates that are then filtered by safety constraints and real-time traffic data.

In summary, while DMs have demonstrated impressive results in offline RL, their extension to online, non-stationary environments is still nascent. Tackling these challenges will require innovations in efficient adaptation mechanisms, exploration-exploitation trade-offs, and robust noise scheduling strategies. Success in this direction could unlock DMs’ potential in safety-critical and continuously evolving real-world applications such as robotics, healthcare decision-making, and autonomous driving.

\subsection{Extending to Multi-Agent and Human-in-the-Loop Systems}
The complexity of multi-agent and human-interactive environments introduces unique challenges for DMs, particularly in terms of coordination, communication, and interpretability. Unlike single-agent settings, where the DM only needs to optimize for one policy, Multi-Agent Reinforcement Learning (MARL) requires the simultaneous modeling of interdependent strategies. Meanwhile, each agent’s action not only influences the environment but also changes the decision-making context of other agents.

For example, in autonomous driving at intersections, multiple self-driving cars must coordinate their maneuvers to avoid collisions while minimizing delays. A naive DM that generates independent trajectories for each car may produce inconsistent behaviors, such as one vehicle yielding while the other simultaneously accelerates. Extending DMs to MARL settings \cite{meng2023offline} requires mechanisms that ensure jointly coherent sampling, so that generated trajectories respect inter-agent dependencies. One promising approach is to combine DMs with communication protocols (e.g., message-passing between agents) to allow agents to exchange intent signals before committing to actions. For instance, a DM could generate both the intended path of a car and an explicit “yield” or “go” communication message, enabling coordination that mirrors human traffic negotiation.

Another frontier is the integration of DMs into human-in-the-loop systems, where agents must adapt their behavior in response to human actions, preferences, or feedback. In collaborative robotics (e.g., factory cobots), a robot arm must anticipate human movements and adjust its trajectory accordingly. A diffusion-based planner could be extended to model human intentions explicitly, for instance by conditioning trajectory sampling on predicted human motion patterns or verbal instructions. Similarly, in assistive healthcare, a DM-driven agent could plan treatment or rehabilitation sequences that account for both the patient’s current state and expected human feedback during the process.

Interpretability is another crucial concern. In multi-agent or human-interactive environments, opaque decision-making can erode trust or lead to unsafe outcomes. For example, if a swarm of drones uses DMs for coordinated search-and-rescue, operators must understand why drones are dispersing in certain patterns. One solution is to couple DMs with intention modeling, where sampled trajectories are mapped to explicit, high-level goals (e.g., “Agent A secures exit route,” “Agent B explores building interior”). This not only facilitates coordination among agents but also improves human interpretability of generated actions.

Furthermore, game-theoretic settings such as pursuit-evasion games highlight the difficulty of multi-agent coordination. In a multi-robot pursuit task, if DMs independently generate chaser trajectories, they may redundantly cover the same area, leaving other regions unguarded. A coordinated DM framework could instead sample complementary strategies, where each robot covers different escape paths of the evader. Such extensions may be inspired by multi-agent planning in POMDPs, where belief-sharing or role assignment reduces redundant exploration.

In summary, extending DMs to multi-agent and human-interactive environments remains an open but promising frontier. Blending diffusion-based generation with communication protocols, intention modeling, and coordination mechanisms can enable scalable solutions. By embedding these elements, DMs can move beyond single-agent optimization toward real-world deployments involving collaborative robots, autonomous vehicles, and human-AI teams.

\subsection{Applying to Large Language Models}
Large Language Models (LLMs) have demonstrated remarkable reasoning and decision-making capabilities in language-based tasks, including question answering, planning, and sequential instruction following. When combined with DMs, these systems can generate multi-modal, temporally structured, and constraint-aware trajectories, enabling richer decision-making policies beyond pure text generation.

For instance, Diffusion-LM \cite{li2022diffusionlm} and DiffuSeq \cite{gong2023diffuseq} illustrate how DMs can generate coherent token sequences under complex future constraints, effectively modeling long-range dependencies in text. DriveDreamer \cite{wang2024drivedreamer} takes this a step further by incorporating LLM-style reasoning into trajectory planning frameworks, improving both interpretability and long-horizon goal satisfaction. For example, in autonomous driving scenarios, the model can reason about multi-step maneuvers—like lane changes, obstacle avoidance, and turns—by planning trajectories that align with high-level navigational goals while respecting safety constraints. This combination of reasoning and generative modeling allows the agent to produce plans that are not only feasible but also explainable, as each step can be traced back to a high-level rationale.

Recent works explore the use of retrieval-augmented prompts or grounding LLM outputs in control domains using diffusion-based methods \cite{chen2023llm, zhu2023diffusionnlp, wei2025automated, zhai2024fine}. For example, a robot could query a dataset of prior trajectories to guide a diffusion-conditioned LLM in generating a novel but feasible manipulation sequence. In automated scientific experiment planning, diffusion-conditioned LLMs can propose sequences of chemical reactions by grounding symbolic actions in prior experimental results, ensuring that generated sequences are physically and chemically plausible. Similarly, in human-robot interaction, retrieval-augmented LLMs can interpret natural language instructions while DMs generate action sequences that respect physical constraints and safety margins.

By combining the reasoning power of LLMs with the stochastic, expressive trajectory modeling of DMs, these approaches enable agents to: 1) Perform long-horizon planning in complex environments, where decisions depend on future consequences; 2) Incorporate multi-modal inputs and constraints, such as text instructions, visual observations, and physical state variables; 3) Enhance interpretability, as each generated trajectory or action sequence can be explained in terms of high-level reasoning steps or retrieved exemplars.

In summary, the integration of LLMs and DMs represents a promising frontier in decision-making AI, bridging symbolic reasoning, natural language understanding, and generative trajectory modeling. Applications range from autonomous driving and robotic manipulation to interactive AI assistants, scientific experiment design, and multi-agent coordination, demonstrating the versatility of this hybrid approach. However, it is crucial and challenging for researchers to enable DMs to handle complex inputs and generate multiple outputs with better performance in LLMs scenarios.

\section{Conclusion}\label{conclusion}
DMs have emerged as a powerful class of generative models that are gaining increasing attention for RL. By leveraging the stochastic denoising process, these models offer a flexible and expressive framework for modeling complex, multi-modal distributions over dynamic environments. Unlike traditional RL and IL, DMs provide a unified approach that integrates learning from data with generative planning capabilities.

This survey has provided a comprehensive overview of the foundations of DMs, including the forward diffusion process, reverse denoising process, and key variants such as DDPMs, DDIMs, and CTDMs. We have discussed how these DMs are applied to the single-agent RL in trajectory optimization, policy learning, IL, exploration augmentation, environmental simulation, and reward modeling, the multi-agent RL in joint trajectory optimization and collaborative policy learning, the online RL, the offline RL, and their emerging theoretical understanding.

We also reviewed a broad spectrum of real-world application scenarios, ranging from robotics and autonomous driving to game playing and healthcare, highlighting the versatility and potential of diffusion-based RL models. Furthermore, we identified key open research issues and future directions, such as sampling efficiency, variance reduction, safety integration, theoretical guarantees, online learning, and LLM integration.

As the field continues to evolve, DMs are expected to play a critical role in bridging the gap between generative modeling and RL modeling. Continued interdisciplinary research across machine learning, control, and real-world systems will be crucial to unlocking their full potential and enabling scalable and robust RL in complex environments.

\begin{acks}
This work was supported in part by grants from the National Natural Science Foundation of China (NSFC) (62172046, 62372047), the Beijing Natural Science Foundation (No. 4232028), the Natural Science Foundation of Guangdong Province (2024A1515011323), Zhuhai Basic and Applied Basic Research Foundation (2220004002619), the Joint Project of Production, Teaching and Research of Zhuhai (2220004002686, 2320004002812), Science and Technology Projects of Social Development in Zhuhai (2320004000213), the Supplemental Funds for Major Scientific Research Projects of Beijing Normal University, Zhuhai (ZHPT2023002), the Fundamental Research Funds for the Central Universities, Higher Education Research Topics of Guangdong Association of Higher Education in the 14th Five-Year Plan under 24GYB207, and the Beijing Normal University Education Reform Project under jx2024139. We also thanks the support from the Interdisciplinary Intelligence Super Computer Center of Beijing Normal University at Zhuhai.
\end{acks}

\bibliographystyle{unsrt}
\bibliography{references}

\begin{thebibliography}{100}

\bibitem{esser2024scaling}
Patrick Esser, Sumith Kulal, Andreas Blattmann, Rahim Entezari, Jonas M{\"u}ller, Harry Saini, Yam Levi, Dominik Lorenz, Axel Sauer, Frederic Boesel, et~al.
\newblock Scaling rectified flow transformers for high-resolution image synthesis.
\newblock In {\em Proceedings of the 41st International Conference on Machine Learning}. JMLR.org, 2024.

\bibitem{diffusion2024platform}
Stable~Diffusion Platform.
\newblock \url{https://stability.ai/news/stable-diffusion-3}.

\bibitem{ho2020denoising}
Jonathan Ho, Ajay Jain, and Pieter Abbeel.
\newblock Denoising diffusion probabilistic models.
\newblock {\em Advances in neural information processing systems}, 33:6840--6851, 2020.

\bibitem{kingma2014auto}
Diederik~P Kingma and Max Welling.
\newblock Auto-encoding variational $\{$Bayes$\}$.
\newblock In {\em International Conference on Learning Representations}, 2014.

\bibitem{goodfellow2014generative}
Ian~J Goodfellow, Jean Pouget-Abadie, Mehdi Mirza, Bing Xu, David Warde-Farley, Sherjil Ozair, Aaron Courville, and Yoshua Bengio.
\newblock Generative adversarial nets.
\newblock {\em Advances in neural information processing systems}, 27, 2014.

\bibitem{lugmayr2022repaint}
Andreas Lugmayr, Martin Danelljan, Andres Romero, Fisher Yu, Radu Timofte, and Luc Van~Gool.
\newblock Repaint: Inpainting using denoising diffusion probabilistic models.
\newblock In {\em Proceedings of the IEEE/CVF conference on computer vision and pattern recognition}, pages 11461--11471, 2022.

\bibitem{austin2021structured}
Jacob Austin, Daniel~D Johnson, Jonathan Ho, Daniel Tarlow, and Rianne Van Den~Berg.
\newblock Structured denoising diffusion models in discrete state-spaces.
\newblock {\em Advances in neural information processing systems}, 34:17981--17993, 2021.

\bibitem{li2022diffusionlm}
Xiang Li, John Thickstun, Ishaan Gulrajani, Percy~S Liang, and Tatsunori~B Hashimoto.
\newblock Diffusion-lm improves controllable text generation.
\newblock {\em Advances in neural information processing systems}, 35:4328--4343, 2022.

\bibitem{lee2021nu}
Junhyeok Lee and Seungu Han.
\newblock Nu-wave: A diffusion probabilistic model for neural audio upsampling.
\newblock In {\em Proc. Interspeech 2021}, pages 1634--1638, 2021.

\bibitem{kong2021diffwave}
Zhifeng Kong, Wei Ping, Jiaji Huang, Kexin Zhao, and Bryan Catanzaro.
\newblock Diffwave: A versatile diffusion model for audio synthesis.
\newblock In {\em International Conference on Learning Representations}, 2021.

\bibitem{ajay2023conditional}
Anurag Ajay, Yilun Du, Abhi Gupta, Joshua Tenenbaum, Tommi Jaakkola, and Pulkit Agrawal.
\newblock Is conditional generative modeling all you need for decision-making?
\newblock In {\em The Eleventh International Conference on Learning Representations}, 2023.

\bibitem{huang2024diffusion}
Renming Huang, Yunqiang Pei, Guoqing Wang, Yangming Zhang, Yang Yang, Peng Wang, and Hengtao Shen.
\newblock Diffusion models as optimizers for efficient planning in offline rl.
\newblock In {\em European Conference on Computer Vision}, pages 1--17. Springer, 2024.

\bibitem{janner2022planning}
Michael Janner, Yilun Du, Joshua Tenenbaum, and Sergey Levine.
\newblock Planning with diffusion for flexible behavior synthesis.
\newblock In {\em International Conference on Machine Learning}, pages 9902--9915. PMLR, 2022.

\bibitem{du2024diffusion}
Hongyang Du, Zonghang Li, Dusit Niyato, Jiawen Kang, Zehui Xiong, Huawei Huang, and Shiwen Mao.
\newblock Diffusion-based reinforcement learning for edge-enabled ai-generated content services.
\newblock {\em IEEE Transactions on Mobile Computing}, 23(9):8902--8918, 2024.

\bibitem{xu2025enhancing}
Changfu Xu, Jianxiong Guo, Yuzhu Liang, Haodong Zou, Jiandian Zeng, Haipeng Dai, Weijia Jia, Jiannong Cao, and Tian Wang.
\newblock Enhancing qoe in collaborative edge systems with feedback diffusion generative scheduling.
\newblock {\em IEEE Transactions on Mobile Computing}, pages 1--16, 2025.

\bibitem{de2022insect}
Guido~CHE de~Croon, JJG Dupeyroux, Sawyer~B Fuller, and James~AR Marshall.
\newblock Insect-inspired ai for autonomous robots.
\newblock {\em Science robotics}, 7(67):eabl6334, 2022.

\bibitem{hu2023planning}
Yihan Hu, Jiazhi Yang, Li~Chen, Keyu Li, Chonghao Sima, Xizhou Zhu, Siqi Chai, Senyao Du, Tianwei Lin, Wenhai Wang, et~al.
\newblock Planning-oriented autonomous driving.
\newblock In {\em Proceedings of the IEEE/CVF conference on computer vision and pattern recognition}, pages 17853--17862, 2023.

\bibitem{liang2024efficient}
Yuzhu Liang, Guo Li, Jianxiong Guo, Qin Liu, Xi~Zheng, and Tian Wang.
\newblock Efficient request scheduling in cross-regional edge collaboration via digital twin networks.
\newblock In {\em 2024 IEEE/ACM 32nd International Symposium on Quality of Service (IWQoS)}, pages 1--6. IEEE, 2024.

\bibitem{mnih2015human}
Volodymyr Mnih, Koray Kavukcuoglu, David Silver, Andrei~A Rusu, Joel Veness, Marc~G Bellemare, Alex Graves, Martin Riedmiller, Andreas~K Fidjeland, Georg Ostrovski, et~al.
\newblock Human-level control through deep reinforcement learning.
\newblock {\em nature}, 518(7540):529--533, 2015.

\bibitem{van2016deep}
Hado Van~Hasselt, Arthur Guez, and David Silver.
\newblock Deep reinforcement learning with double q-learning.
\newblock In {\em Proceedings of the AAAI conference on artificial intelligence}, volume~30, 2016.

\bibitem{wang2016dueling}
Ziyu Wang, Tom Schaul, Matteo Hessel, Hado Hasselt, Marc Lanctot, and Nando Freitas.
\newblock Dueling network architectures for deep reinforcement learning.
\newblock In {\em International conference on machine learning}, pages 1995--2003. PMLR, 2016.

\bibitem{schulman2015trust}
John Schulman, Sergey Levine, Pieter Abbeel, Michael Jordan, and Philipp Moritz.
\newblock Trust region policy optimization.
\newblock In {\em International conference on machine learning}, pages 1889--1897. PMLR, 2015.

\bibitem{schulman2017proximal}
John Schulman, Filip Wolski, Prafulla Dhariwal, Alec Radford, and Oleg Klimov.
\newblock Proximal policy optimization algorithms.
\newblock {\em arXiv preprint arXiv:1707.06347}, 2017.

\bibitem{silver2014deterministic}
David Silver, Guy Lever, Nicolas Heess, Thomas Degris, Daan Wierstra, and Martin Riedmiller.
\newblock Deterministic policy gradient algorithms.
\newblock In {\em International conference on machine learning}, pages 387--395. Pmlr, 2014.

\bibitem{lillicrap2015continuous}
Timothy~P Lillicrap, Jonathan~J Hunt, Alexander Pritzel, Nicolas Heess, Tom Erez, Yuval Tassa, David Silver, and Daan Wierstra.
\newblock Continuous control with deep reinforcement learning.
\newblock {\em arXiv preprint arXiv:1509.02971}, 2015.

\bibitem{fujimoto2018addressing}
Scott Fujimoto, Herke Hoof, and David Meger.
\newblock Addressing function approximation error in actor-critic methods.
\newblock In {\em International conference on machine learning}, pages 1587--1596. PMLR, 2018.

\bibitem{haarnoja2018soft}
Tuomas Haarnoja, Aurick Zhou, Pieter Abbeel, and Sergey Levine.
\newblock Soft actor-critic: Off-policy maximum entropy deep reinforcement learning with a stochastic actor.
\newblock In {\em International conference on machine learning}, pages 1861--1870. Pmlr, 2018.

\bibitem{zhu2023diffusionrl}
Zhengbang Zhu, Hanye Zhao, Haoran He, Yichao Zhong, Shenyu Zhang, Haoquan Guo, Tingting Chen, and Weinan Zhang.
\newblock Diffusion models for reinforcement learning: A survey.
\newblock {\em arXiv preprint arXiv:2311.01223}, 2023.

\bibitem{fujimoto2019benchmarking}
Scott Fujimoto, Edoardo Conti, Mohammad Ghavamzadeh, and Joelle Pineau.
\newblock Benchmarking batch deep reinforcement learning algorithms.
\newblock {\em arXiv preprint arXiv:1910.01708}, 2019.

\bibitem{lu2023contrastive}
Cheng Lu, Huayu Chen, Jianfei Chen, Hang Su, Chongxuan Li, and Jun Zhu.
\newblock Contrastive energy prediction for exact energy-guided diffusion sampling in offline reinforcement learning.
\newblock In {\em International Conference on Machine Learning}, pages 22825--22855. PMLR, 2023.

\bibitem{xu2024phd}
Changfu Xu.
\newblock Phd forum abstract: Diffusion-based task scheduling for efficient ai-generated content in edge networks.
\newblock In {\em 2024 23rd ACM/IEEE International Conference on Information Processing in Sensor Networks (IPSN)}, pages 333--334. IEEE, 2024.

\bibitem{wang2023diffusion}
Zhendong Wang, Jonathan~J Hunt, and Mingyuan Zhou.
\newblock Diffusion policies as an expressive policy class for offline reinforcement learning.
\newblock In {\em The Eleventh International Conference on Learning Representations}, 2023.

\bibitem{lu2023synthetic}
Cong Lu, Philip Ball, Yee~Whye Teh, and Jack Parker-Holder.
\newblock Synthetic experience replay.
\newblock {\em Advances in Neural Information Processing Systems}, 36:46323--46344, 2023.

\bibitem{venkatraman2023reasoning}
Siddarth Venkatraman, Shivesh Khaitan, Ravi~Tej Akella, John Dolan, Jeff Schneider, and Glen Berseth.
\newblock Reasoning with latent diffusion in offline reinforcement learning.
\newblock In {\em The Twelfth International Conference on Learning Representations}, 2023.

\bibitem{he2023diffusion}
Haoran He, Chenjia Bai, Kang Xu, Zhuoran Yang, Weinan Zhang, Dong Wang, Bin Zhao, and Xuelong Li.
\newblock Diffusion model is an effective planner and data synthesizer for multi-task reinforcement learning.
\newblock {\em Advances in neural information processing systems}, 36:64896--64917, 2023.

\bibitem{hegde2023generating}
Shashank Hegde, Sumeet Batra, KR~Zentner, and Gaurav Sukhatme.
\newblock Generating behaviorally diverse policies with latent diffusion models.
\newblock {\em Advances in Neural Information Processing Systems}, 36:7541--7554, 2023.

\bibitem{carvalho2023motion}
Joao Carvalho, An~T Le, Mark Baierl, Dorothea Koert, and Jan Peters.
\newblock Motion planning diffusion: Learning and planning of robot motions with diffusion models.
\newblock In {\em 2023 IEEE/RSJ International Conference on Intelligent Robots and Systems (IROS)}, pages 1916--1923. IEEE, 2023.

\bibitem{kang2023efficient}
Bingyi Kang, Xiao Ma, Chao Du, Tianyu Pang, and Shuicheng Yan.
\newblock Efficient diffusion policies for offline reinforcement learning.
\newblock {\em Advances in Neural Information Processing Systems}, 36:67195--67212, 2023.

\bibitem{cao2024survey}
Hanqun Cao, Cheng Tan, Zhangyang Gao, Yilun Xu, Guangyong Chen, Pheng-Ann Heng, and Stan~Z Li.
\newblock A survey on generative diffusion models.
\newblock {\em IEEE Transactions on Knowledge and Data Engineering}, 2024.

\bibitem{yang2023diffusion}
Ling Yang, Zhilong Zhang, Yang Song, Shenda Hong, Runsheng Xu, Yue Zhao, Wentao Zhang, Bin Cui, and Ming-Hsuan Yang.
\newblock Diffusion models: A comprehensive survey of methods and applications.
\newblock {\em ACM Computing Surveys}, 56(4):1--39, 2023.

\bibitem{croitoru2023diffusion}
Florinel-Alin Croitoru, Vlad Hondru, Radu~Tudor Ionescu, and Mubarak Shah.
\newblock Diffusion models in vision: A survey.
\newblock {\em IEEE Transactions on Pattern Analysis and Machine Intelligence}, 45(9):10850--10869, 2023.

\bibitem{zhang2023text}
Chenshuang Zhang, Chaoning Zhang, Mengchun Zhang, and In~So Kweon.
\newblock Text-to-image diffusion models in generative ai: A survey.
\newblock {\em arXiv preprint arXiv:2303.07909}, 2023.

\bibitem{ulhaq2022efficient}
Anwaar Ulhaq and Naveed Akhtar.
\newblock Efficient diffusion models for vision: A survey.
\newblock {\em arXiv preprint arXiv:2210.09292}, 2022.

\bibitem{zhu2023diffusionnlp}
Yuansong Zhu and Yu~Zhao.
\newblock Diffusion models in nlp: A survey.
\newblock {\em arXiv preprint arXiv:2303.07576}, 2023.

\bibitem{zou2023survey}
Hao Zou, Zae~Myung Kim, and Dongyeop Kang.
\newblock A survey of diffusion models in natural language processing.
\newblock {\em arXiv preprint arXiv:2305.14671}, 2023.

\bibitem{li2023diffusion}
Yifan Li, Kun Zhou, Wayne~Xin Zhao, and Ji-Rong Wen.
\newblock Diffusion models for non-autoregressive text generation: a survey.
\newblock In {\em Proceedings of the Thirty-Second International Joint Conference on Artificial Intelligence}, pages 6692--6701, 2023.

\bibitem{zhang2023survey}
Chenshuang Zhang, Chaoning Zhang, Sheng Zheng, Mengchun Zhang, Maryam Qamar, Sung-Ho Bae, and In~So Kweon.
\newblock A survey on audio diffusion models: Text to speech synthesis and enhancement in generative ai.
\newblock {\em arXiv preprint arXiv:2303.13336}, 2023.

\bibitem{guo2023diffusion}
Zhiye Guo, Jian Liu, Yanli Wang, Mengrui Chen, Duolin Wang, Dong Xu, and Jianlin Cheng.
\newblock Diffusion models in bioinformatics: A new wave of deep learning revolution in action.
\newblock {\em arXiv preprint arXiv:2302.10907}, 2023.

\bibitem{du2024enhancing}
Hongyang Du, Ruichen Zhang, Yinqiu Liu, Jiacheng Wang, Yijing Lin, Zonghang Li, Dusit Niyato, Jiawen Kang, Zehui Xiong, Shuguang Cui, et~al.
\newblock Enhancing deep reinforcement learning: A tutorial on generative diffusion models in network optimization.
\newblock {\em IEEE Communications Surveys \& Tutorials}, 2024.

\bibitem{bellman1957dp}
Richard Bellman.
\newblock A markovian decision process.
\newblock {\em Journal of Mathematics and Mechanics}, pages 679--684, 1957.

\bibitem{puterman2014mdp}
Martin~L Puterman.
\newblock {\em Markov Decision Processes: Discrete Stochastic Dynamic Programming}.
\newblock John Wiley \& Sons, 2014.

\bibitem{kaelbling1998pomdp}
Leslie~Pack Kaelbling, Michael~L Littman, and Anthony~R Cassandra.
\newblock Planning and acting in partially observable stochastic domains.
\newblock {\em Artificial Intelligence}, 101(1--2):99--134, 1998.

\bibitem{moerland2023model}
Thomas~M Moerland, Joost Broekens, Aske Plaat, Catholijn~M Jonker, et~al.
\newblock Model-based reinforcement learning: A survey.
\newblock {\em Foundations and Trends{\textregistered} in Machine Learning}, 16(1):1--118, 2023.

\bibitem{watkins1992q}
Christopher~JCH Watkins and Peter Dayan.
\newblock Q-learning.
\newblock {\em Machine learning}, 8:279--292, 1992.

\bibitem{sutton1999policy}
Richard~S Sutton, David McAllester, Satinder Singh, and Yishay Mansour.
\newblock Policy gradient methods for reinforcement learning with function approximation.
\newblock {\em Advances in neural information processing systems}, 12, 1999.

\bibitem{konda1999actor}
Vijay Konda and John Tsitsiklis.
\newblock Actor-critic algorithms.
\newblock {\em Advances in neural information processing systems}, 12, 1999.

\bibitem{tan1993multi}
Ming Tan.
\newblock Multi-agent reinforcement learning: Independent vs. cooperative agents.
\newblock In {\em Proceedings of the tenth international conference on machine learning}, pages 330--337, 1993.

\bibitem{lowe2017multi}
Ryan Lowe, Yi~I Wu, Aviv Tamar, Jean Harb, OpenAI Pieter~Abbeel, and Igor Mordatch.
\newblock Multi-agent actor-critic for mixed cooperative-competitive environments.
\newblock {\em Advances in neural information processing systems}, 30, 2017.

\bibitem{sunehag2017value}
Peter Sunehag, Guy Lever, Audrunas Gruslys, Wojciech~Marian Czarnecki, Vinicius Zambaldi, Max Jaderberg, Marc Lanctot, Nicolas Sonnerat, Joel~Z Leibo, Karl Tuyls, et~al.
\newblock Value-decomposition networks for cooperative multi-agent learning.
\newblock {\em arXiv preprint arXiv:1706.05296}, 2017.

\bibitem{rashid2018qmix}
Tabish Rashid, Mikayel Samvelyan, Christian Schroeder, Gregory Farquhar, Jakob Foerster, and Shimon Whiteson.
\newblock Qmix: Monotonic value function factorisation for deep multi-agent reinforcement learning.
\newblock In {\em International Conference on Machine Learning}, pages 4295--4304. PMLR, 2018.

\bibitem{yu2022surprising}
Chao Yu, Akash Velu, Eugene Vinitsky, Jiaxuan Gao, Yu~Wang, Alexandre Bayen, and Yi~Wu.
\newblock The surprising effectiveness of ppo in cooperative multi-agent games.
\newblock {\em Advances in neural information processing systems}, 35:24611--24624, 2022.

\bibitem{papoudakis2020benchmarking}
Georgios Papoudakis, Filippos Christianos, Lukas Sch{\"a}fer, and Stefano~V Albrecht.
\newblock Benchmarking multi-agent deep reinforcement learning algorithms in cooperative tasks.
\newblock {\em arXiv preprint arXiv:2006.07869}, 2020.

\bibitem{he2024hierarchical}
Shunfan He, Ronghao Zheng, Senlin Zhang, and Meiqin Liu.
\newblock Hierarchical policy optimization for cooperative multi-agent reinforcement learning.
\newblock In {\em 2024 IEEE International Conference on Systems, Man, and Cybernetics (SMC)}, pages 5087--5092. IEEE, 2024.

\bibitem{sohl2015deep}
Jascha Sohl-Dickstein, Eric Weiss, Niru Maheswaranathan, and Surya Ganguli.
\newblock Deep unsupervised learning using nonequilibrium thermodynamics.
\newblock In {\em International conference on machine learning}, pages 2256--2265. pmlr, 2015.

\bibitem{van2017neural}
Aaron Van Den~Oord, Oriol Vinyals, et~al.
\newblock Neural discrete representation learning.
\newblock {\em Advances in neural information processing systems}, 30, 2017.

\bibitem{rombach2022high}
Robin Rombach, Andreas Blattmann, Dominik Lorenz, Patrick Esser, and Bj{\"o}rn Ommer.
\newblock High-resolution image synthesis with latent diffusion models.
\newblock In {\em Proceedings of the IEEE/CVF conference on computer vision and pattern recognition}, pages 10684--10695, 2022.

\bibitem{song2021denoising}
Jiaming Song, Chenlin Meng, and Stefano Ermon.
\newblock Denoising diffusion implicit models.
\newblock In {\em International Conference on Learning Representations}, 2021.

\bibitem{song2021score}
Yang Song, Jascha Sohl-Dickstein, Diederik~P Kingma, Abhishek Kumar, Stefano Ermon, and Ben Poole.
\newblock Score-based generative modeling through stochastic differential equations.
\newblock In {\em International Conference on Learning Representations}, 2021.

\bibitem{lu2022dpm}
Cheng Lu, Yuhao Zhou, Fan Bao, Jianfei Chen, Chongxuan Li, and Jun Zhu.
\newblock Dpm-solver: A fast ode solver for diffusion probabilistic model sampling in around 10 steps.
\newblock {\em Advances in Neural Information Processing Systems}, 35:5775--5787, 2022.

\bibitem{dhariwal2021diffusion}
Prafulla Dhariwal and Alexander Nichol.
\newblock Diffusion models beat gans on image synthesis.
\newblock {\em Advances in neural information processing systems}, 34:8780--8794, 2021.

\bibitem{yao2025enhancing}
Zhi Yao, Zhiqing Tang, Wenmian Yang, and Weijia Jia.
\newblock Enhancing llm qos through cloud-edge collaboration: A diffusion-based multi-agent reinforcement learning approach.
\newblock {\em IEEE Transactions on Services Computing}, 18(3):1412--1427, 2025.

\bibitem{li2023hierarchical}
Wenhao Li, Xiangfeng Wang, Bo~Jin, and Hongyuan Zha.
\newblock Hierarchical diffusion for offline decision making.
\newblock In {\em International Conference on Machine Learning}, pages 20035--20064. PMLR, 2023.

\bibitem{wu2024diffusion}
Zixuan Wu, Sean Ye, Manisha Natarajan, and Matthew~C Gombolay.
\newblock Diffusion-reinforcement learning hierarchical motion planning in adversarial multi-agent games.
\newblock {\em arXiv preprint arXiv:2403.10794}, 2024.

\bibitem{li2023beyond}
Zhuoran Li, Ling Pan, and Longbo Huang.
\newblock Beyond conservatism: Diffusion policies in offline multi-agent reinforcement learning.
\newblock {\em arXiv preprint arXiv:2307.01472}, 2023.

\bibitem{qi2024diffusion}
Xinyue Qi, Jianhang Tang, Jiangming Jin, and Yang Zhang.
\newblock Diffusion-based multi-agent reinforcement learning with communication.
\newblock In {\em 2024 IEEE VTS Asia Pacific Wireless Communications Symposium (APWCS)}, pages 1--6. IEEE, 2024.

\bibitem{xu2024beyond}
Zhiwei Xu, Hangyu Mao, Nianmin Zhang, Xin Xin, Pengjie Ren, Dapeng Li, Bin Zhang, Guoliang Fan, Zhumin Chen, Changwei Wang, et~al.
\newblock Beyond local views: Global state inference with diffusion models for cooperative multi-agent reinforcement learning.
\newblock {\em arXiv preprint arXiv:2408.09501}, 2024.

\bibitem{jiang2023motiondiffuser}
Chiyu Jiang, Andre Cornman, Cheolho Park, Benjamin Sapp, Yin Zhou, Dragomir Anguelov, et~al.
\newblock Motiondiffuser: Controllable multi-agent motion prediction using diffusion.
\newblock In {\em Proceedings of the IEEE/CVF conference on computer vision and pattern recognition}, pages 9644--9653, 2023.

\bibitem{zhu2024madiff}
Zhengbang Zhu, Minghuan Liu, Liyuan Mao, Bingyi Kang, Minkai Xu, Yong Yu, Stefano Ermon, and Weinan Zhang.
\newblock Madiff: Offline multi-agent learning with diffusion models.
\newblock {\em Advances in Neural Information Processing Systems}, 37:4177--4206, 2024.

\bibitem{li2025dof}
Chao Li, Ziwei Deng, Chenxing Lin, Wenqi Chen, Yongquan Fu, Weiquan Liu, Chenglu Wen, Cheng Wang, and Siqi Shen.
\newblock Dof: A diffusion factorization framework for offline multi-agent reinforcement learning.
\newblock In {\em The Thirteenth International Conference on Learning Representations}, 2025.

\bibitem{wang2023diffusionRM}
Wenjie Wang, Yiyan Xu, Fuli Feng, Xinyu Lin, Xiangnan He, and Tat-Seng Chua.
\newblock Diffusion recommender model.
\newblock In {\em Proceedings of the 46th International ACM SIGIR Conference on Research and Development in Information Retrieval}, pages 832--841, 2023.

\bibitem{yang2023generate}
Zhengyi Yang, Jiancan Wu, Zhicai Wang, Xiang Wang, Yancheng Yuan, and Xiangnan He.
\newblock Generate what you prefer: Reshaping sequential recommendation via guided diffusion.
\newblock {\em Advances in Neural Information Processing Systems}, 36:24247--24261, 2023.

\bibitem{gong2023diffuseq}
Shansan Gong, Mukai Li, Jiangtao Feng, Zhiyong Wu, and Lingpeng Kong.
\newblock Diffuseq: Sequence to sequence text generation with diffusion models.
\newblock In {\em The Eleventh International Conference on Learning Representations}, 2023.

\bibitem{zhang2024offline}
Jiazhi Zhang, Yuhu Cheng, Shuo Cao, and Xuesong Wang.
\newblock Offline reinforcement learning with reverse diffusion guide policy.
\newblock {\em IEEE Transactions on Industrial Informatics}, 2024.

\bibitem{he2023diffcps}
Longxiang He, Li~Shen, Linrui Zhang, Junbo Tan, and Xueqian Wang.
\newblock Diffcps: Diffusion model based constrained policy search for offline reinforcement learning.
\newblock {\em arXiv preprint arXiv:2310.05333}, 2023.

\bibitem{hu2023instructed}
Jifeng Hu, Yanchao Sun, Sili Huang, SiYuan Guo, Hechang Chen, Li~Shen, Lichao Sun, Yi~Chang, and Dacheng Tao.
\newblock Instructed diffuser with temporal condition guidance for offline reinforcement learning.
\newblock {\em arXiv preprint arXiv:2306.04875}, 2023.

\bibitem{zhang2024motiondiffuse}
Mingyuan Zhang, Zhongang Cai, Liang Pan, Fangzhou Hong, Xinying Guo, Lei Yang, and Ziwei Liu.
\newblock Motiondiffuse: Text-driven human motion generation with diffusion model.
\newblock {\em IEEE transactions on pattern analysis and machine intelligence}, 46(6):4115--4128, 2024.

\bibitem{wang2025diffad}
Tao Wang, Cong Zhang, Xingguang Qu, Kun Li, Weiwei Liu, and Chang Huang.
\newblock Diffad: A unified diffusion modeling approach for autonomous driving.
\newblock {\em arXiv preprint arXiv:2503.12170}, 2025.

\bibitem{chen2023polydiffuse}
Jiacheng Chen, Ruizhi Deng, and Yasutaka Furukawa.
\newblock Polydiffuse: Polygonal shape reconstruction via guided set diffusion models.
\newblock {\em Advances in Neural Information Processing Systems}, 36:1863--1888, 2023.

\bibitem{liao2025diffusiondrive}
Bencheng Liao, Shaoyu Chen, Haoran Yin, Bo~Jiang, Cheng Wang, Sixu Yan, Xinbang Zhang, Xiangyu Li, Ying Zhang, Qian Zhang, et~al.
\newblock Diffusiondrive: Truncated diffusion model for end-to-end autonomous driving.
\newblock In {\em Proceedings of the Computer Vision and Pattern Recognition Conference}, pages 12037--12047, 2025.

\bibitem{liang2023adaptdiffuser}
Zhixuan Liang, Yao Mu, Mingyu Ding, Fei Ni, Masayoshi Tomizuka, and Ping Luo.
\newblock Adaptdiffuser: diffusion models as adaptive self-evolving planners.
\newblock In {\em Proceedings of the 40th International Conference on Machine Learning}, pages 20725--20745, 2023.

\bibitem{song2022learning}
Haoran Song, Di~Luan, Wenchao Ding, Michael~Y Wang, and Qifeng Chen.
\newblock Learning to predict vehicle trajectories with model-based planning.
\newblock In {\em Conference on Robot Learning}, pages 1035--1045. PMLR, 2022.

\bibitem{xu2024accelerating}
Changfu Xu, Jianxiong Guo, Wanyu Lin, Haodong Zou, Wentao Fan, Tian Wang, Xiaowen Chu, and Jiannong Cao.
\newblock Accelerating aigc services with latent action diffusion scheduling in edge networks.
\newblock {\em arXiv preprint arXiv:2412.18212}, 2024.

\bibitem{hansen2023idql}
Philippe Hansen-Estruch, Ilya Kostrikov, Michael Janner, Jakub~Grudzien Kuba, and Sergey Levine.
\newblock Idql: Implicit q-learning as an actor-critic method with diffusion policies.
\newblock {\em arXiv preprint arXiv:2304.10573}, 2023.

\bibitem{dhariwal2021classifier}
Prafulla Dhariwal and Alex Nichol.
\newblock Diffusion models beat gans on image synthesis.
\newblock In {\em Advances in Neural Information Processing Systems (NeurIPS)}, 2021.

\bibitem{ho2016generative}
Jonathan Ho and Stefano Ermon.
\newblock Generative adversarial imitation learning.
\newblock {\em Advances in neural information processing systems}, 29, 2016.

\bibitem{jang2022bc}
Eric Jang, Alex Irpan, Mohi Khansari, Daniel Kappler, Frederik Ebert, Corey Lynch, Sergey Levine, and Chelsea Finn.
\newblock Bc-z: Zero-shot task generalization with robotic imitation learning.
\newblock In {\em Conference on Robot Learning}, pages 991--1002. PMLR, 2022.

\bibitem{ross2011dagger}
Stéphane Ross, Geoff Gordon, and Drew Bagnell.
\newblock A reduction of imitation learning and structured prediction to no-regret online learning.
\newblock In {\em AISTATS}, pages 627--635, 2011.

\bibitem{abbeel2004apprenticeship}
Pieter Abbeel and Andrew~Y Ng.
\newblock Apprenticeship learning via inverse reinforcement learning.
\newblock In {\em Proceedings of the twenty-first international conference on Machine learning}, page~1, 2004.

\bibitem{ziebart2008maximum}
Brian~D Ziebart, Andrew~L Maas, J~Andrew Bagnell, Anind~K Dey, et~al.
\newblock Maximum entropy inverse reinforcement learning.
\newblock In {\em Aaai}, volume~8, pages 1433--1438. Chicago, IL, USA, 2008.

\bibitem{xie2025latent}
Amber Xie, Oleh Rybkin, Dorsa Sadigh, and Chelsea Finn.
\newblock Latent diffusion planning for imitation learning.
\newblock {\em arXiv preprint arXiv:2504.16925}, 2025.

\bibitem{liu2024diff}
Xiao Liu, Yifan Zhou, Fabian Weigend, Shubham Sonawani, Shuhei Ikemoto, and Heni~Ben Amor.
\newblock Diff-control: A stateful diffusion-based policy for imitation learning.
\newblock In {\em 2024 IEEE/RSJ International Conference on Intelligent Robots and Systems (IROS)}, pages 7453--7460. IEEE, 2024.

\bibitem{mnih2016asynchronous}
Volodymyr Mnih, Adria~Puigdomenech Badia, Mehdi Mirza, Alex Graves, Timothy Lillicrap, Tim Harley, David Silver, and Koray Kavukcuoglu.
\newblock Asynchronous methods for deep reinforcement learning.
\newblock In {\em International conference on machine learning}, pages 1928--1937. PmLR, 2016.

\bibitem{pathak2017curiosity}
Deepak Pathak, Pulkit Agrawal, Alexei~A Efros, and Trevor Darrell.
\newblock Curiosity-driven exploration by self-supervised prediction.
\newblock In {\em International conference on machine learning}, pages 2778--2787. PMLR, 2017.

\bibitem{burda2018exploration}
Yuri Burda, Harrison Edwards, Amos Storkey, and Oleg Klimov.
\newblock Exploration by random network distillation.
\newblock {\em arXiv preprint arXiv:1810.12894}, 2018.

\bibitem{song2023consistency}
Yang Song, Prafulla Dhariwal, Mark Chen, and Ilya Sutskever.
\newblock Consistency models.
\newblock In {\em International Conference on Machine Learning}, pages 32211--32252. PMLR, 2023.

\bibitem{janner2019trust}
Michael Janner, Justin Fu, Marvin Zhang, and Sergey Levine.
\newblock When to trust your model: Model-based policy optimization.
\newblock {\em Advances in neural information processing systems}, 32, 2019.

\bibitem{zhou2024redi}
Shiyang Zhou, Zehao Gu, Yun Xiong, Yang Luo, Qiang Wang, and Xiaofeng Gao.
\newblock Redi: Recurrent diffusion model for probabilistic time series forecasting.
\newblock In {\em Proceedings of the 33rd ACM International Conference on Information and Knowledge Management}, pages 3505--3514, 2024.

\bibitem{wang2024drivedreamer}
Xiaofeng Wang, Zheng Zhu, Guan Huang, Xinze Chen, Jiagang Zhu, and Jiwen Lu.
\newblock Drivedreamer: Towards real-world-drive world models for autonomous driving.
\newblock In {\em European Conference on Computer Vision}, pages 55--72. Springer, 2024.

\bibitem{ni2023metadiffuser}
Fei Ni, Jianye Hao, Yao Mu, Yifu Yuan, Yan Zheng, Bin Wang, and Zhixuan Liang.
\newblock Metadiffuser: Diffusion model as conditional planner for offline meta-rl.
\newblock In {\em International Conference on Machine Learning}, pages 26087--26105. PMLR, 2023.

\bibitem{wang2024safe}
Ruhan Wang and Dongruo Zhou.
\newblock Safe decision transformer with learning-based constraints.
\newblock In {\em Neurips Safe Generative AI Workshop}, 2024.

\bibitem{du2021survey}
Wei Du and Shifei Ding.
\newblock A survey on multi-agent deep reinforcement learning: from the perspective of challenges and applications.
\newblock {\em Artificial Intelligence Review}, 54(5):3215--3238, 2021.

\bibitem{pateria2021hierarchical}
Shubham Pateria, Budhitama Subagdja, Ah-hwee Tan, and Chai Quek.
\newblock Hierarchical reinforcement learning: A comprehensive survey.
\newblock {\em ACM Computing Surveys (CSUR)}, 54(5):1--35, 2021.

\bibitem{luo2024text}
Calvin Luo, Mandy He, Zilai Zeng, and Chen Sun.
\newblock Text-aware diffusion for policy learning.
\newblock {\em Advances in Neural Information Processing Systems}, 37:46226--46253, 2024.

\bibitem{kim2024robust}
Woo~Kyung Kim, Minjong Yoo, and Honguk Woo.
\newblock Robust policy learning via offline skill diffusion.
\newblock In {\em Proceedings of the AAAI Conference on Artificial Intelligence}, volume~38, pages 13177--13184, 2024.

\bibitem{liang2024skilldiffuser}
Zhixuan Liang, Yao Mu, Hengbo Ma, Masayoshi Tomizuka, Mingyu Ding, and Ping Luo.
\newblock Skilldiffuser: Interpretable hierarchical planning via skill abstractions in diffusion-based task execution.
\newblock In {\em Proceedings of the IEEE/CVF Conference on Computer Vision and Pattern Recognition}, pages 16467--16476, 2024.

\bibitem{sun2025score}
Mingyang Sun, Pengxiang Ding, Weinan Zhang, and Donglin Wang.
\newblock Score-based diffusion policy compatible with reinforcement learning via optimal transport.
\newblock {\em arXiv preprint arXiv:2502.12631}, 2025.

\bibitem{kong2021fast}
Zhifeng Kong and Wei Ping.
\newblock On fast sampling of diffusion probabilistic models.
\newblock In {\em ICML Workshop on Invertible Neural Networks, Normalizing Flows, and Explicit Likelihood Models}, 2023.

\bibitem{lin2024survey}
Jianghao Lin, Jiaqi Liu, Jiachen Zhu, Yunjia Xi, Chengkai Liu, Yangtian Zhang, Yong Yu, and Weinan Zhang.
\newblock A survey on diffusion models for recommender systems.
\newblock {\em arXiv preprint arXiv:2409.05033}, 2024.

\bibitem{wolf2025diffusion}
Rosa Wolf, Yitian Shi, Sheng Liu, and Rania Rayyes.
\newblock Diffusion models for robotic manipulation: A survey.
\newblock {\em arXiv preprint arXiv:2504.08438}, 2025.

\bibitem{chi2023diffusion}
Cheng Chi, Zhenjia Xu, Siyuan Feng, Eric Cousineau, Yilun Du, Benjamin Burchfiel, Russ Tedrake, and Shuran Song.
\newblock Diffusion policy: Visuomotor policy learning via action diffusion.
\newblock {\em The International Journal of Robotics Research}, 2023.

\bibitem{liu2025diffusion}
Zhiang Liu, Yang Liu, and Yongchun Fang.
\newblock Diffusion model-based path follower for a salamander-like robot.
\newblock {\em IEEE Transactions on Neural Networks and Learning Systems}, 2025.

\bibitem{deo2018convolutional}
Nachiket Deo and Mohan~M Trivedi.
\newblock Convolutional social pooling for vehicle trajectory prediction.
\newblock In {\em Proceedings of the IEEE conference on computer vision and pattern recognition workshops}, pages 1468--1476, 2018.

\bibitem{westny2023mtp}
Theodor Westny, Joel Oskarsson, Bj{\"o}rn Olofsson, and Erik Frisk.
\newblock Mtp-go: Graph-based probabilistic multi-agent trajectory prediction with neural odes.
\newblock {\em IEEE Transactions on Intelligent Vehicles}, 8(9):4223--4236, 2023.

\bibitem{brehmer2023edgi}
Johann Brehmer, Joey Bose, Pim De~Haan, and Taco~S Cohen.
\newblock Edgi: Equivariant diffusion for planning with embodied agents.
\newblock {\em Advances in Neural Information Processing Systems}, 36:63818--63834, 2023.

\bibitem{he2022diffusionbert}
Zhengfu He, Tianxiang Sun, Kuanning Wang, Xuanjing Huang, and Xipeng Qiu.
\newblock Diffusionbert: Improving generative masked language models with diffusion models.
\newblock {\em arXiv preprint arXiv:2211.15029}, 2022.

\bibitem{nachmani2021zero}
Eliya Nachmani and Shaked Dovrat.
\newblock Zero-shot translation using diffusion models.
\newblock {\em arXiv preprint arXiv:2111.01471}, 2021.

\bibitem{xu2024dynamic}
Changfu Xu, Jianxiong Guo, Yupeng Li, Haodong Zou, Weijia Jia, and Tian Wang.
\newblock Dynamic parallel multi-server selection and allocation in collaborative edge computing.
\newblock {\em IEEE Transactions on Mobile Computing}, pages 1--15, 2024.

\bibitem{liang2025collaborative}
Yuzhu Liang, Mujun Yin, Wenhua Wang, Qin Liu, Liang Wang, Xi~Zheng, and Tian Wang.
\newblock Collaborative edge server placement for maximizing qos with distributed data cleaning.
\newblock {\em IEEE Transactions on Services Computing}, 2025.

\bibitem{zou2024fine}
Haodong Zou, Jianxiong Guo, Jiandian Zeng, Yupeng Li, Jiannong Cao, and Tian Wang.
\newblock Fine-grained service lifetime optimization for energy-constrained edge-edge collaboration.
\newblock In {\em 2024 IEEE 44th International Conference on Distributed Computing Systems (ICDCS)}, pages 565--576. IEEE, 2024.

\bibitem{xu2024incorporating}
Changfu Xu, Jianxiong Guo, Jiandian Zeng, Yupeng Li, Jiannong Cao, and Tian Wang.
\newblock Incorporating startup delay into collaborative edge computing for superior task efficiency.
\newblock In {\em 2024 IEEE/ACM 32nd International Symposium on Quality of Service (IWQoS)}, pages 1--10. IEEE, 2024.

\bibitem{zhang2025diffusion}
Zheng Zhang, Jingjing Wang, Jianrui Chen, Hang Fu, Ziheng Tong, and Chunxiao Jiang.
\newblock Diffusion-based reinforcement learning for cooperative offloading and resource allocation in multi-uav assisted edge-enabled metaverse.
\newblock {\em IEEE Transactions on Vehicular Technology}, pages 1--13, 2025.

\bibitem{ma2024multimodal}
Haokai Ma, Yimeng Yang, Lei Meng, Ruobing Xie, and Xiangxu Meng.
\newblock Multimodal conditioned diffusion model for recommendation.
\newblock In {\em Companion Proceedings of the ACM Web Conference 2024}, pages 1733--1740, 2024.

\bibitem{wang2024leadrec}
Weidong Wang, Yan Tang, and Kun Tian.
\newblock Leadrec: Towards personalized sequential recommendation via guided diffusion.
\newblock In {\em International Conference on Intelligent Computing}, pages 3--15. Springer, 2024.

\bibitem{blattmann2022retrieval}
Andreas Blattmann, Robin Rombach, Kaan Oktay, Jonas M{\"u}ller, and Bj{\"o}rn Ommer.
\newblock Retrieval-augmented diffusion models.
\newblock {\em Advances in Neural Information Processing Systems}, 35:15309--15324, 2022.

\bibitem{valevski2024diffusion}
Dani Valevski, Yaniv Leviathan, Moab Arar, and Shlomi Fruchter.
\newblock Diffusion models are real-time game engines.
\newblock {\em arXiv preprint arXiv:2408.14837}, 2024.

\bibitem{alonso2024diffusion}
Eloi Alonso, Adam Jelley, Vincent Micheli, Anssi Kanervisto, Amos~J Storkey, Tim Pearce, and Fran{\c{c}}ois Fleuret.
\newblock Diffusion for world modeling: Visual details matter in atari.
\newblock {\em Advances in Neural Information Processing Systems}, 37:58757--58791, 2024.

\bibitem{kazerouni2023diffusion}
Amirhossein Kazerouni, Ehsan~Khodapanah Aghdam, Moein Heidari, Reza Azad, Mohsen Fayyaz, Ilker Hacihaliloglu, and Dorit Merhof.
\newblock Diffusion models in medical imaging: A comprehensive survey.
\newblock {\em Medical image analysis}, 88:102846, 2023.

\bibitem{al2024diffusion}
Abdullah al~Nomaan~Nafi, Md~Alamgir~Hossain, Rakib Hossain~Rifat, Md~Mahabub~Uz Zaman, Md~Manjurul~Ahsan, and Shivakumar Raman.
\newblock Diffusion-based approaches in medical image generation and analysis.
\newblock {\em arXiv e-prints}, pages arXiv--2412, 2024.

\bibitem{geng2023diffusion}
Jinkun Geng, Xiubo Liang, Hongzhi Wang, and Yu~Zhao.
\newblock Diffusion policies as multi-agent reinforcement learning strategies.
\newblock In {\em International Conference on Artificial Neural Networks}, pages 356--364. Springer, 2023.

\bibitem{sattarov2023findiff}
Timur Sattarov, Marco Schreyer, and Damian Borth.
\newblock Findiff: Diffusion models for financial tabular data generation.
\newblock In {\em Proceedings of the Fourth ACM International Conference on AI in Finance}, pages 64--72, 2023.

\bibitem{lin2024energydiff}
Nan Lin, Peter Palensky, and Pedro~P Vergara.
\newblock Energydiff: Universal time-series energy data generation using diffusion models.
\newblock {\em arXiv preprint arXiv:2407.13538}, 2024.

\bibitem{geng2025diffusion}
Zeyi Geng, Linfeng Yang, and Wuqing Yu.
\newblock A diffusion model-based framework to enhance the robustness of non-intrusive load disaggregation.
\newblock {\em Energy}, 320:135423, 2025.

\bibitem{midjourney}
Midjourney Platform.
\newblock \url{https://www.midjourney.com/home}.

\bibitem{huggingface}
Hugging~Face Platform.
\newblock \url{https://huggingface.co/spaces}.

\bibitem{lyu2022accelerating}
Zhaoyang Lyu, Xudong Xu, Ceyuan Yang, Dahua Lin, and Bo~Dai.
\newblock Accelerating diffusion models via early stop of the diffusion process.
\newblock {\em arXiv preprint arXiv:2205.12524}, 2022.

\bibitem{igl2018deep}
Maximilian Igl, Luisa Zintgraf, Tuan~Anh Le, Frank Wood, and Shimon Whiteson.
\newblock Deep variational reinforcement learning for pomdps.
\newblock In {\em International conference on machine learning}, pages 2117--2126. PMLR, 2018.

\bibitem{xu2022prompting}
Mengdi Xu, Yikang Shen, Shun Zhang, Yuchen Lu, Ding Zhao, Joshua Tenenbaum, and Chuang Gan.
\newblock Prompting decision transformer for few-shot policy generalization.
\newblock In {\em International Conference on Machine Learning}, pages 24631--24645. PMLR, 2022.

\bibitem{gulcehre2020rl}
Caglar Gulcehre, Ziyu Wang, Alexander Novikov, Thomas Paine, Sergio G{\'o}mez, Konrad Zolna, Rishabh Agarwal, Josh~S Merel, Daniel~J Mankowitz, Cosmin Paduraru, et~al.
\newblock Rl unplugged: A suite of benchmarks for offline reinforcement learning.
\newblock {\em Advances in Neural Information Processing Systems}, 33:7248--7259, 2020.

\bibitem{meng2023offline}
Linghui Meng, Muning Wen, Chenyang Le, Xiyun Li, Dengpeng Xing, Weinan Zhang, Ying Wen, Haifeng Zhang, Jun Wang, Yaodong Yang, et~al.
\newblock Offline pre-trained multi-agent decision transformer.
\newblock {\em Machine Intelligence Research}, 20(2):233--248, 2023.

\bibitem{wei2025automated}
Dixiao Wei, Peng Yi, Jinlong Lei, Yiguang Hong, and Yuchuan Du.
\newblock An automated reinforcement learning reward design framework with large language model for cooperative platoon coordination.
\newblock {\em arXiv preprint arXiv:2504.19480}, 2025.

\bibitem{chen2023llm}
Siwei Chen, Anxing Xiao, and David Hsu.
\newblock Llm-state: Open world state representation for long-horizon task planning with large language model.
\newblock {\em arXiv preprint arXiv:2311.17406}, 2023.

\bibitem{zhai2024fine}
Simon Zhai, Hao Bai, Zipeng Lin, Jiayi Pan, Peter Tong, Yifei Zhou, Alane Suhr, Saining Xie, Yann LeCun, Yi~Ma, et~al.
\newblock Fine-tuning large vision-language models as decision-making agents via reinforcement learning.
\newblock {\em Advances in neural information processing systems}, 37:110935--110971, 2024.

\end{thebibliography}


\end{document}